%% file: relational.tex
\documentclass[12pt]{article}

\usepackage{bm}
\usepackage{amsfonts}
\usepackage{amsmath}
\usepackage{amssymb}

\usepackage{hyperref} 

\usepackage{graphicx} 
\usepackage{color} 

\usepackage{helvet}

\usepackage{geometry}
\title{Learning to relate images: Mapping units, \\complex cells and simultaneous eigenspaces}
\author{Roland Memisevic\\
University of Frankfurt\\
ro@cs.uni-frankfurt.de
}

\date{\today}

\begin{document}

\maketitle

\begin{abstract} 
A fundamental operation in many vision tasks, including motion understanding, stereopsis, 
visual odometry, or invariant recognition,  
is \emph{establishing correspondences} between images or between images and data from other modalities. 
We present an analysis of the role that multiplicative interactions play in learning such 
correspondences, and we show how learning and inferring relationships between images can be viewed as 
detecting rotations in the eigenspaces shared among a set of orthogonal matrices. 
We review a variety of recent multiplicative sparse coding methods in light of this observation. 
We also review how the squaring operation performed by 
energy models and by models of complex cells can be thought of as a way to implement
multiplicative interactions. This suggests that the main utility of including complex cells in 
computational models of vision may be that they can encode \emph{relations} not invariances. 
\end{abstract}

\section{Introduction}

\emph{Correspondence} is arguably the most ubiquitous computational primitive in vision: 
{\bf Tracking} amounts to establishing correspondences between frames; 
{\bf stereo vision} between different views of a scene; 
{\bf optical flow} between any two images; 
{\bf invariant recognition} between images and invariant descriptions in memory; 
{\bf odometry} between images and motion information; 
{\bf action recognition} between frames; etc.
In these and many other tasks, the relationship \emph{between} images not the content 
of a single image carries the relevant information. 
Representing structures within a single image, such as contours, can be also considered 
as an instance of a correspondence problem, namely between areas, or pixels, within an 
image\footnote{The importance of image correspondence in action understanding is nicely 
illustrated in Heider and Simmel's 1944 video of geometric objects engaged in 
various ``social activities'' \cite{heidersimmel}
(althouth the original intent of that video goes beyond making a case for correspondences). 
Each single frame depicts a rather meaningless set of geometric objects and conveys almost 
no information about the content of the movie. The only way to understand the movie is 
by understanding the motions and actions, and thus by decoding the relationships 
\emph{between} frames.}.
The fact that correspondence is such a common operation across vision suggests that 
the task of \emph{representing relations} may have to be kept in mind when trying to 
build autonomous vision systems and when trying to understand biological vision.  

A lot of progress has been made recently in building models that learn to solve 
tasks like object recognition from independent, static images. 
One of the reasons for the recent progress is the use of \emph{local features},  
which help virtually eliminate the notoriously difficult problems of occlusions 
and small invariances. 
A central finding is that the right choice of features not the choice of 
high-level classifier or computational pipeline are what typically 
makes a system work well. 
Interestingly, some of the best performing recognition models are highly biologically consistent, 
in that they are based on features that are learned unsupervised from data. 
Besides being biological plausible, feature learning comes with various benefits, 
such as helping overcome tedious engineering, 
helping adapt to new domains 
and allowing for some degree of \emph{end-to-end learning}
in place of constructing, and then combining, a large number of modules to solve a task.
The fact that tasks like object recognition can be solved using biologically consistent, 
learning based methods raises the question whether understanding \emph{relations} 
can be amenable to learning in the same way. If so, this may open up the road to learning 
based and/or biologically consistent approaches to a much larger variety of problems than 
static object recognition, and perhaps also beyond vision. 

\begin{figure}
\begin{center}
\resizebox{4.0cm}{!}{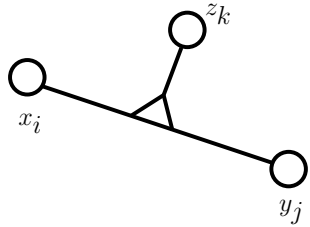} 
\end{center}
\caption{Symbolic representation of a \emph{mapping unit} \cite{MappingsHinton1981}. 
The triangle symbolizes multiplicative interactions between the three variables
$z_k$, $x_i$ and $y_j$. The value of any one of the three variables 
is a function of the product of all the others. 
(\cite{MappingsHinton1981}).}
\label{figure:mapping}
\end{figure}

In this paper, we review a variety of recent methods that address correspondence tasks 
by learning local features. 
We discuss how the common computational principle behind all these methods 
are \emph{multiplicative interactions}, which were introduced to the vision 
community $30$ years ago under the terms ``mapping units'' \cite{MappingsHinton1981} and 
``dynamic mappings'' \cite{MappingsMalsburg1981}. 
An illustration of mapping units is shown in Figure \ref{figure:mapping}: 
The three variables shown in the figure interact multiplicatively, and as a result, 
each variable (say, $z_k$) can be thought of as dynamically \emph{modulating} 
the connections between other variables in the model ($x_i$ and $y_j$). 
Likewise, the value of any variable (eg., $y_j$) can be thought of as 
depending on the product of the other variables ($x_i$, $z_k$) \cite{MappingsHinton1981}.  
This is in contrast to common feature learning models like ICA, 
Restricted Boltzmann Machines, auto-encoder networks and many others, 
all of which are based on bi-partite networks, that do not involve any three-way
multiplicative interactions. 
In these models, independent hidden variables interact with independent observable 
variables, such that the value of any variable depends on a weighted \emph{sum} 
not product of the other variables. 
Closely related to models of mapping units are energy models (for example, \cite{adelson1985spatiotemporal}), 
which may be thought of as a way to ``emulate'' multiplicative interactions by computing squares. 

We shall show how both mapping units and energy models can be viewed as ways to learn 
and detect rotations in a set of \emph{shared invariant subspaces of a set of commuting matrices}. 
Our analysis may help understand why action recognition methods seem to profit from squaring 
non-linearities (for example, \cite{201106-cvpr-le}), and it predicts that squaring and cross-products 
will be helpful, in general, in applications that involve representing relations. 

\subsection{A brief history of multiplicative interactions}
\label{section:briefhistory}
Shortly after mapping units were introduced in $1981$, energy models \cite{adelson1985spatiotemporal} 
received a lot of attention. 
Energy models are closely related to cross-correlation models \cite{arndt1995human}, 
which, in turn, are a type of multiplicative interaction model. 
Energy models have been used as a way to model motion (relating time frames in 
a video) \cite{adelson1985spatiotemporal} and 
stereo vision (relating images across different eyes or cameras) \cite{ODF}. 
An energy model is a computational unit that relates images by summing over  
squared responses of, typically two, linear projections of input data. This 
operation can be shown to encode translations independently of 
content \cite{FleetBinocular}, \cite{QianStereoAndMotion} (cf. Section \ref{section:factorization}). 

Early approaches to building and applying energy and cross-correlation models 
were based entirely on hand-wiring (see, for example, \cite{QianStereoAndMotion},
\cite{sangerStereoGabor}, \cite{FleetBinocular}). 
Practically all of these models use Gabor filters as the linear receptive fields 
whose responses are squared and summed. 
The focus on Gabor features has somewhat biased the analysis of energy models 
to focus on the \emph{Fourier-spectrum} 
as the main object of interest (see, for example, \cite{FleetBinocular, QianStereoAndMotion}).
As we shall discuss in Section \ref{section:factorization}, Fourier-components arise just as 
the special case of one transformation class, namely \emph{translation}, and many of the analyses 
apply more generally and to other types of transformation. 

Gabor-based energy models have also been applied monocularly. In this case they encode features 
independently of the Fourier-phase of the input. 
As a result, their responses are invariant to small translations as well as  
to contrast variations of the input. 
In part for this reason, energy models have been popular in models of 
\emph{complex cells}, which are known to show similar invariance properties (see, 
for example, \cite{NaturalImageStatistics}). 

Shortly after energy and cross-correlation models emerged, 
there has been some attention on learning invariances with higher-order neural networks, 
which are neural networks trained on polynomial basis expansions of their inputs, 
\cite{Giles_higherorder}. 
Higher-order neural networks can be composed of units that compute sums of products. These units 
are sometimes referred to 
as ``Sigma-Pi-units'' \cite{rumelhart86general} (where ``Pi'' stands for product and ``Sigma'' for sum).  
\cite{smolensky:tensor}, at about the same time,  
discussed how multiplicative interactions make it possible to build distributed 
representations of symbolic data. 

In 1995, Kohonen introduced the 
``Adaptive Subspace Self-Organizing Map'' (ASSOM) \cite{Kohonen1995}, 
which computes sums over squared filter responses to represent data. 
Like the energy model, the ASSOM is based on the idea that the sum of 
squared responses is invariant to various properties of its inputs.  
In contrast to the early energy models, the ASSOM is trained from data. 
Inspired by the ASSOM, \cite{HyvarinenISA} introduced ``Independent Subspace Analysis'' (ISA), 
which puts the same idea into the context of more conventional sparse coding models. 
Extensions of this work are topographic ICA \cite{HyvarinenISA} 
and \cite{Welling02}, where sums are computed not over separate but over shared 
groups of squared filter responses.  

In a parallel line of work, bi-linear models were used as an approach to 
learning in the presence of multiplicative interactions \cite{tenenbaum00separating}. 
This early work on bi-linear models used these as global models trained on whole images 
rather than using local receptive fields. 
In contrast to more recent approaches to learning with 
multiplicative interactions, training typically involved filling a two-dimensional 
grid with data that shows two types of variability (sometimes called ``style'' and ``content''). 
The purpose of bi-linear models is then to untangle the two degrees of freedom in the 
data. 
More recent work does not make this distinction, and the purpose 
of multiplicative hidden variables is merely to capture the multiple ways in which 
two images can be related. \cite{GrimesRao}, \cite{olshausenetal}, \cite{imtrans}, 
for example, show how multiplicative interactions make it possible to model 
the multitude of relationships between frames in natural videos. 
\cite{imtrans} also show how they allow us to model more general classes 
of relations between images. 
An earlier multiplicative interaction model, that is also related to bi-linear models, 
is the ``routing-circuit'' \cite{olshausen_thesis}.  

Multiplicative interactions have also been used to model structure within static images, 
which can be thought of as modeling higher-order relations, and, in particular, 
pair-wise products, between pixel intensities 
(for example, \cite{Karklin-Lewicki-06-NIPS, HyvarinenISA, gaussianscalemixture, hoyerContours, mcrbm, spikeandslabrbm_aistats, higherordergradientbased}).

Recently, \cite{gsm} showed how multiplicative interactions between a \emph{class-label} 
and a feature vector can be viewed as an invariant classifier, where each class is 
represented by a manifold of allowable transformations. 
This work may be viewed as a modern version of the model that introduced the term 
mapping units in 1981 \cite{MappingsHinton1981}. The main difference between 2011 and 1981
is that models are now trained from large datasets.

\section{Learning to relate images}
\label{section:learningfeaturesthatencoderelations}

\subsection{Feature learning}
\label{section:featurelearning}
We briefly review standard feature learning models in this section and we discuss relational 
feature learning in Section \ref{section:modelingrelations}. We discuss extensions of relational 
models and how they relate to complex cells and to energy models in Section \ref{section:factorization}.

\begin{figure}
    \begin{center}
    \begin{tabular}{cc}
        \resizebox{5.0cm}{!}{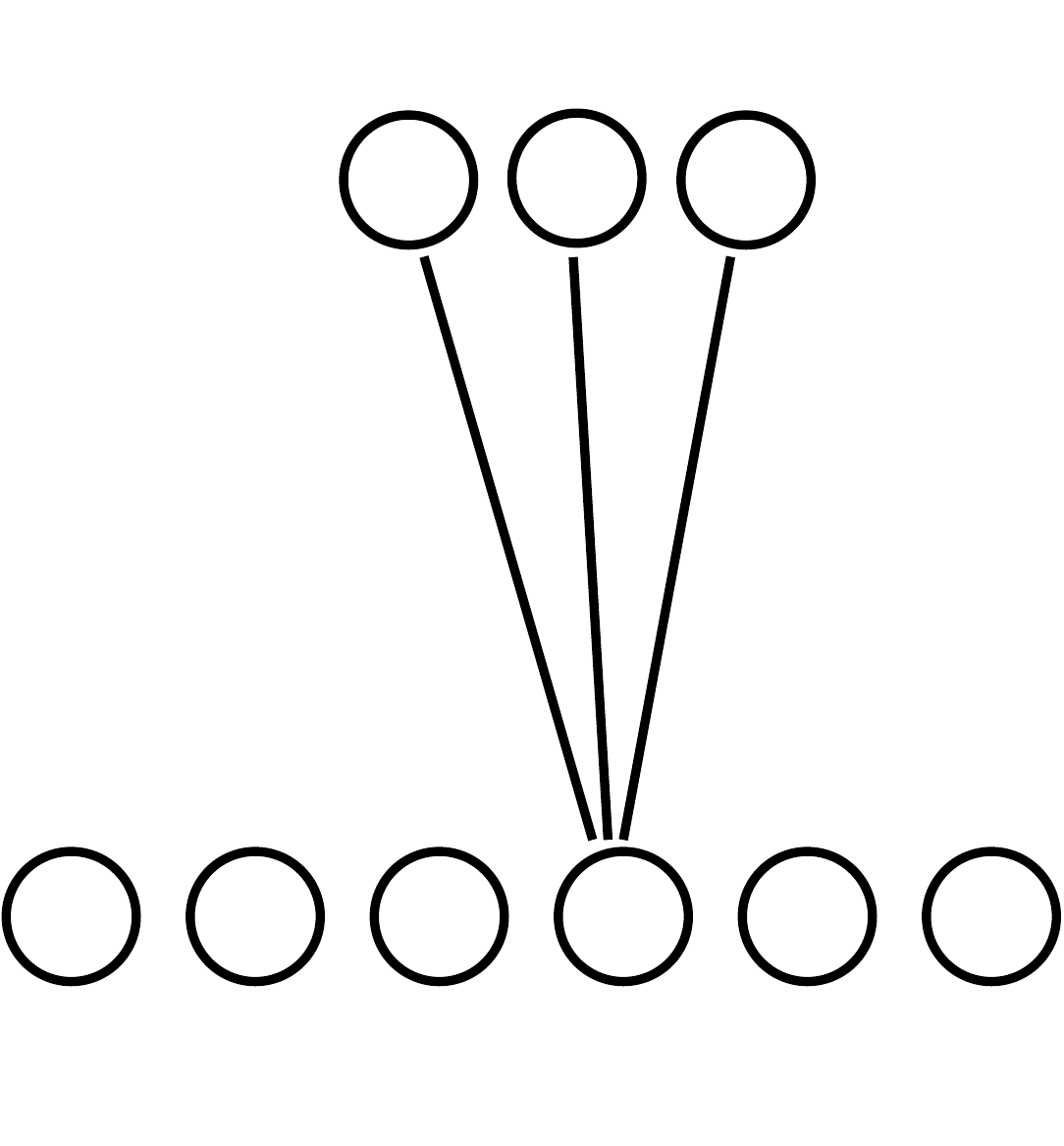} \hspace{30pt} &
        \resizebox{5.5cm}{!}{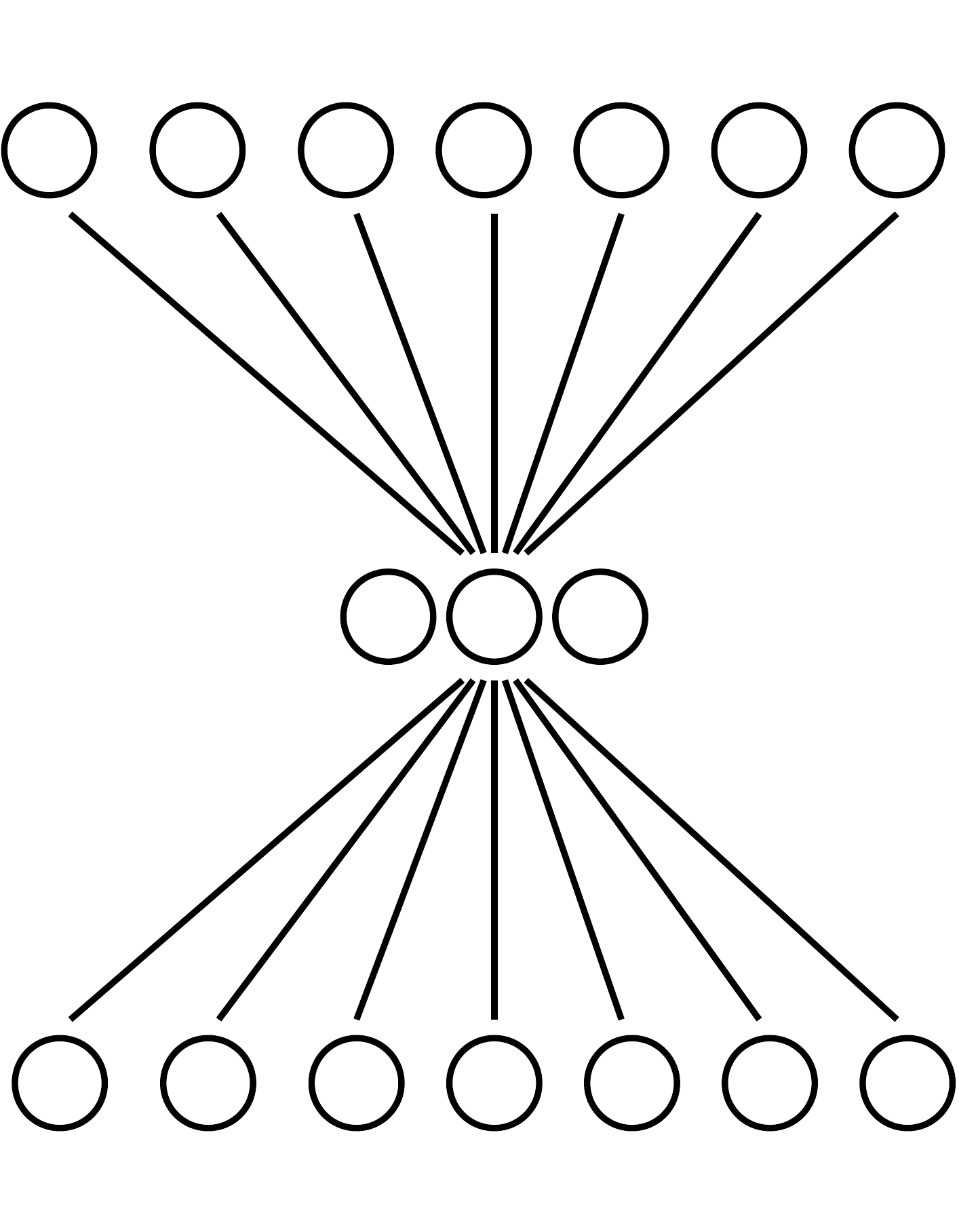} \\
        (a) & (b) 
    \end{tabular}
    \end{center}
\label{figure:sparsecodingmodel}
\caption{(a) Sparse coding graphical model. (b) Auto-encoder network.}
\end{figure}

Practically all standard feature learning models can be represented by a graphical model 
like the one shown in Figure \ref{figure:sparsecodingmodel} (a).
The model is a bi-partite network that connects a set of unobserved, latent 
variables $z_k$ with a set of observable variables (for example, pixels) $y_j$.
The weights $w_{jk}$, which connect pixel $y_j$ with hidden unit $z_k$,
are learned from a set of training 
images $\{\bm{y}^\alpha\}_{\alpha=1, \ldots, N}$. 
The vector of latent variables $\bm{z}=(z_k)_{k=1\ldots K}$ in Figure \ref{figure:sparsecodingmodel} (a)
is considered to be unobserved, so one has to infer it, separately for each training case, along 
with the model parameters for training. 
The graphical model shown in the figure represents how the dependencies between components 
$y_i$ and $z_k$ are parameterized, but it does not define a model or learning algorithm. 
A large variety of models and learning algorithms can be parameterized as in the figure, 
including principal components, mixture models, k-means clustering, 
or restricted Boltzmann machines \cite{cd}. 
Each of these can in principle be used as a feature learning method (see, for example, 
\cite{coatessinglelayer} for a recent quantitative comparison). 

For the hidden variables to extract useful structure from the images, their 
capacity needs to be constrained. The simplest form of constraining it is to 
let the dimensionality $K$ be smaller than the dimensionality $J$ of the images. 
Learning in this case amounts to performing dimensionality reduction. 
It has become obvious recently that it is more useful in most applications to
use an \emph{over-complete} representation, that is, $K>J$, and to constrain the 
capacity of the latent variables instead by forcing the hidden unit activities to 
be \emph{sparse}. In Figure \ref{figure:sparsecodingmodel}, and in what follows, we 
use $K<J$ to symbolize the fact that $\bm{z}$ is capacity-constrained, 
but it should be kept in mind that capacity can be (and often is) constrained in other ways.
The most common operations in the model, after training, are:
``Inference'' (or ``Analysis''): Given image $\bm{y}$, compute $\bm{z}$; and 
``Generation'' (or ``Synthesis''): Invent a latent vector $\bm{z}$, then compute $\bm{y}$.

A simple way to train a model, given training images, is by minimizing reconstruction 
error combined with a sparsity encouraging term for the hidden 
variables (for example, \cite{OlshausenField_nature}):
\begin{equation}
\sum_\alpha \big(\| \bm{y}^\alpha - \sum_k z_k^\alpha W_{.k} \|^2 + \lambda |z_k^\alpha|\big)
\end{equation}
Optimization is with respect to both $W=(w_{jk})_{j=1\ldots J, k=1\ldots K}$ and
all $\bm{z}^\alpha$. 
For this end, it is common to alternate between optimizing $W$ and optimizing all $\bm{z}^\alpha$.
After training, inference then amounts to minimizing the same expression for test images 
(with $W$ fixed). 

To avoid iterative optimization during inference, one can eliminate $\bm{z}$ by 
defining it implicitly as a function of $\bm{y}$. 
A common choice of function is $\bm{z}=\sigma\left( A\bm{y} \right)$ where 
$A$ is a matrix and $\sigma(\cdot)$ is a squashing non-linearity, such as 
$\sigma(a)=(1+\exp(-a))^{-1}$, which confines the values of $\bm{z}$ to reside 
in a fixed interval. 
This model is the well-known auto-encoder (for example, \cite{denoisingAE}) and it is 
depicted in Figure \ref{figure:sparsecodingmodel}. 
Learning amounts to minimizing reconstruction error with respect to both $A$ and $W$.  
In practice, it is common to enforce $A:=W^\mathrm{T}$ in order to reduce the number 
of parameters and for consistency with other sparse coding models. 

One can add a penalty term that encourages sparsity of the latent variables.
Alternatively, one can train auto-encoders, such that they de-noise 
corrupted version of their inputs, which can be achieved by simply feeding in 
corrupted inputs during training (but measuring reconstruction error with respect to 
the original data). This turns auto-encoders into ``de-noising auto-encoders'' \cite{denoisingAE}, 
which show properties similar to other sparse coding methods, but inference, 
like in a standard auto-encoder, is a simple feed-forward mapping. 

A technique similar to the auto-encoder is the Restricted Boltzmann machine (RBM): 
RBMs define the joint probability distribution 
\begin{equation}
\label{eq:rbm}
p(\bm{y}, \bm{z})=\frac{1}{Z}\exp\big(\sum_{jk}w_{jk}y_jz_k\big),
\end{equation} 
from which one can derive 
\begin{equation}
p(z_k|\bm{y})=\mathrm{sigmoid}\big(\sum_j w_{jk}y_j\big)
\quad \mbox{and} \quad
p(y_j|\bm{z})=\mathrm{sigmoid}\big(\sum_j w_{jk}z_k\big),
\end{equation}
showing that inference, again, amounts to a linear mapping plus non-linearity. 
Learning amounts to maximizing the average log-probability $\frac{1}{N}\sum_\alpha \log p(\bm{y}^\alpha)$ 
of the training data. Since the derivatives with respect to the parameters are not tractable
(due to the normalizing constant $Z$ in Eq. \ref{eq:rbm}), 
it is common to use approximate Gibbs sampling in order to approximate them. 
This leads to a Hebbian-like learning rule known as contrastive divergence training \cite{cd}. 

Another common sparse coding method is {\bf independent components analysis} (ICA) 
(for example, \cite{NaturalImageStatistics}). 
One way to train an ICA-model that is complete (that is, where $\bm{z}$ has the same size as $\bm{y}$) 
is by encouraging latent responses to be sparse, while preventing weights from becoming 
degenerate \cite{NaturalImageStatistics}:
\begin{eqnarray}
    &&\min_W \|W^\mathrm{T} \bm{y} \|_1 \\
    &&\mathrm{s.t.} \quad W^\mathrm{T}W=I
\end{eqnarray}
Enforcing the constraint can be inefficient in practice, since it requires an eigen decomposition.  

For most feature learning models, inference and generation are variations of the two linear mappings: 
\begin{equation}
z_k = \sum_j w_{jk} y_j
\label{eq:inference}
\end{equation}
\begin{equation}
y_j = \sum_k w_{jk} z_k
\label{eq:generation}
\end{equation}
The set of model parameters $W_{\cdot k}$ for any $k$ are typically referred to 
as ``features'' or ``filters'' (although a more appropriate term would be ``basis functions'';
we shall use these interchangeably). 
Practically all methods yield Gabor-like features when trained on natural images. 
An advantage of non-linear models, such as RBM's and auto-encoders, is that 
stacking them makes it possible to learn feature hierarchies (``deep learning'') \cite{deepbelief}. 

In practice, it is common to add bias terms, such that inference and generation 
(Eqs. \ref{eq:inference} and \ref{eq:generation}) are affine not linear functions, 
for example, $y_j = \sum_k w_{jk} z_k + b_j$ for some parameter $b_j$. 
We shall refrain from adding bias terms to avoid clutter, noting that, alternatively, 
one may think of $\bm{y}$ and $\bm{z}$ as being in ``homogeneous'' coordinates, containing 
an extra, constant $1$-dimension.  

Feature learning is typically performed on small images patches of size between 
around $5\times 5$ and $50\times 50$ pixels. 
One reason for this is that training and inference can be computationally demanding. 
More important, local features make it possible to deal with images of different 
size, and to deal with occlusions and local object variations. 
Given a trained model, two common ways to perform invariant recognition on test 
images are: 

{\bf ``Bag-Of-Features'':} Crop patches around interest points (such as SIFT or Harris corners),  
compute latent representation $\bm{z}$ for each patch, collapse (add up) all representations 
to obtain a single vector $\bm{z}^\mathrm{Image}$, classify $\bm{z}^\mathrm{Image}$ using a
standard classifier. 
There are several variations of this scheme, including using an extra clustering-step before 
collapsing features, or using a histogram-similarity in place of Euclidean distance for the collapsed 
representation. 

{\bf ``Convolutional'':} Crop patches from the image along a regular grid; compute $\bm{z}$ for 
each patch; concatenate all descriptors into a very large vector $\bm{z}^\mathrm{Image}$; 
classify $\bm{z}^\mathrm{Image}$ using a standard classifier. 
One can also use combinations of the two schemes (see, for example \cite{coatessinglelayer}). 

Local features yield highly competitive performance in object recognition tasks 
(for example, \cite{coatessinglelayer}). In the next section we discuss recent 
approaches to extending feature learning to encode relations between, as opposed to 
content within, images.

\subsection{Encoding relations}
\label{section:modelingrelations}
\begin{figure}
    \begin{center}
        \resizebox{7.0cm}{!}{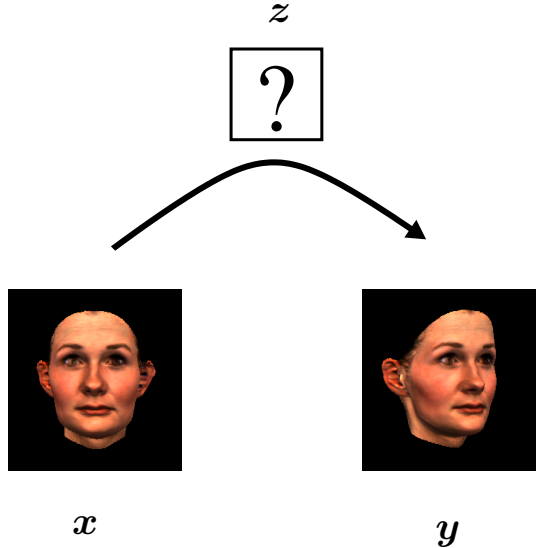}
    \end{center}
\caption{Learning to encode relations: We consider the task of learning latent variables $\bm{z}$
that encode the relationship between images $\bm{x}$ and $\bm{y}$, independently of their content.}
\label{figure:relationalfeaturelearning}
\end{figure}

We now consider the task of learning relations between two images $\bm{x}$ and $\bm{y}$ as 
illustrated\footnote{Face images taken from the data-base described in \cite{mpifaces}} in Figure \ref{figure:relationalfeaturelearning}, 
and we discuss the role of multiplicative interactions when learning relations.

\subsubsection{The need for multiplicative interactions}
A naive approach to modeling relations between two images would be to 
perform sparse coding on the \emph{concatenation}. 
A hidden unit in such a model would receive as input the sum 
of two projections, one from each image. 
To detect a particular transformation, the two receptive fields would need to be 
defined, such that one receptive field is the other modified by the transformation 
that the hidden unit is supposed to detect. 
The net input that the hidden unit receives will then tend to be high 
for image pairs showing the transformation. 
However, the net input will equally dependent on the images \emph{themselves}. 
The reason is that hidden variables are akin to logical ``OR''-gates, 
which accumulate evidence (see, for example \cite{zetzsche2005} for a discussion). 

\begin{figure}
\begin{center}
\begin{tabular}{p{130pt}p{130pt}p{100pt}}
    \resizebox{4.5cm}{!}{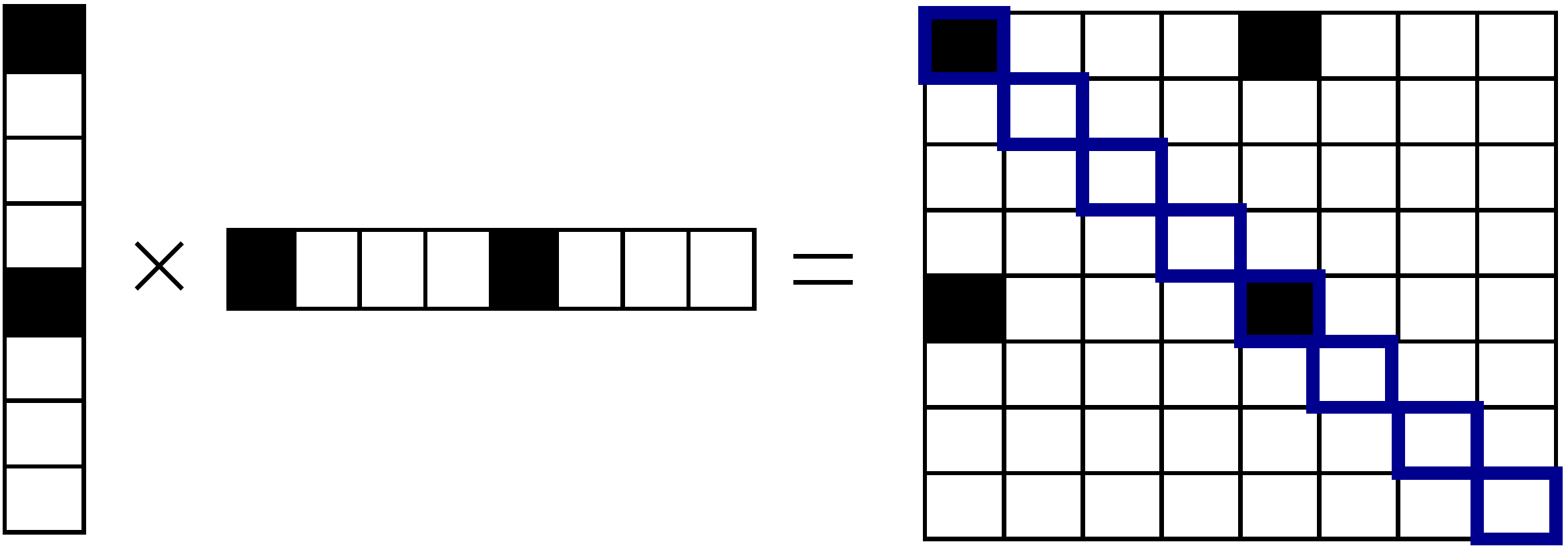} &
    \resizebox{4.5cm}{!}{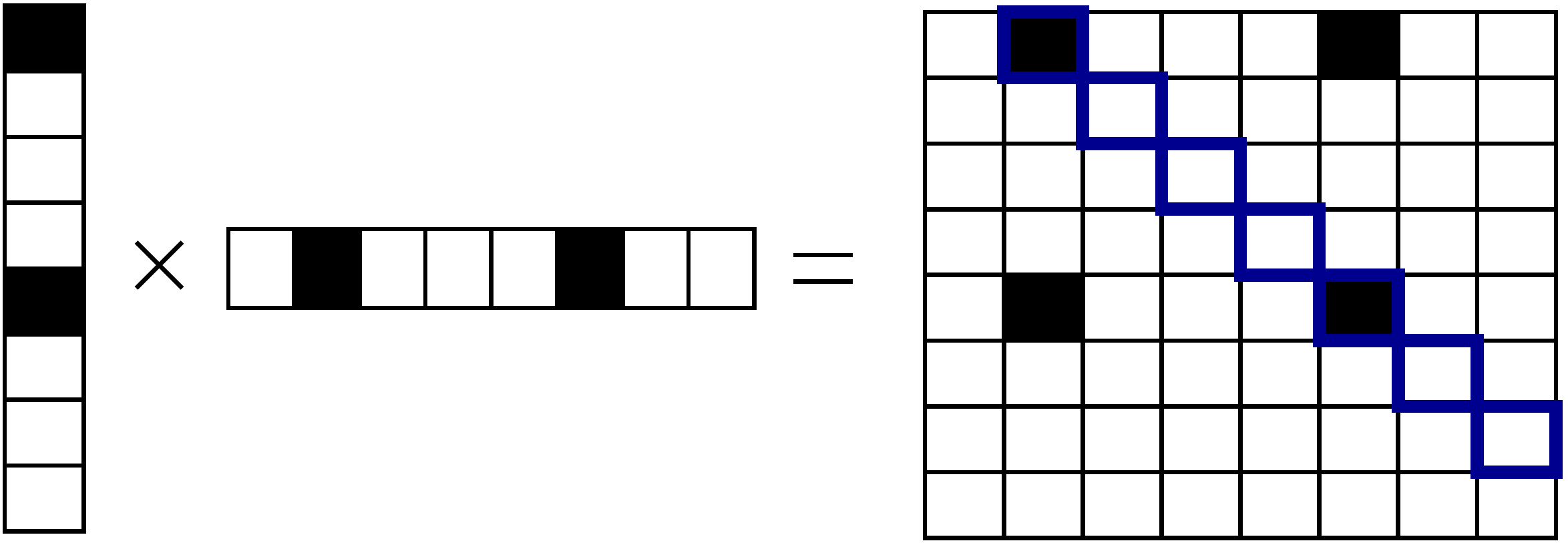} &
    \resizebox{4.5cm}{!}{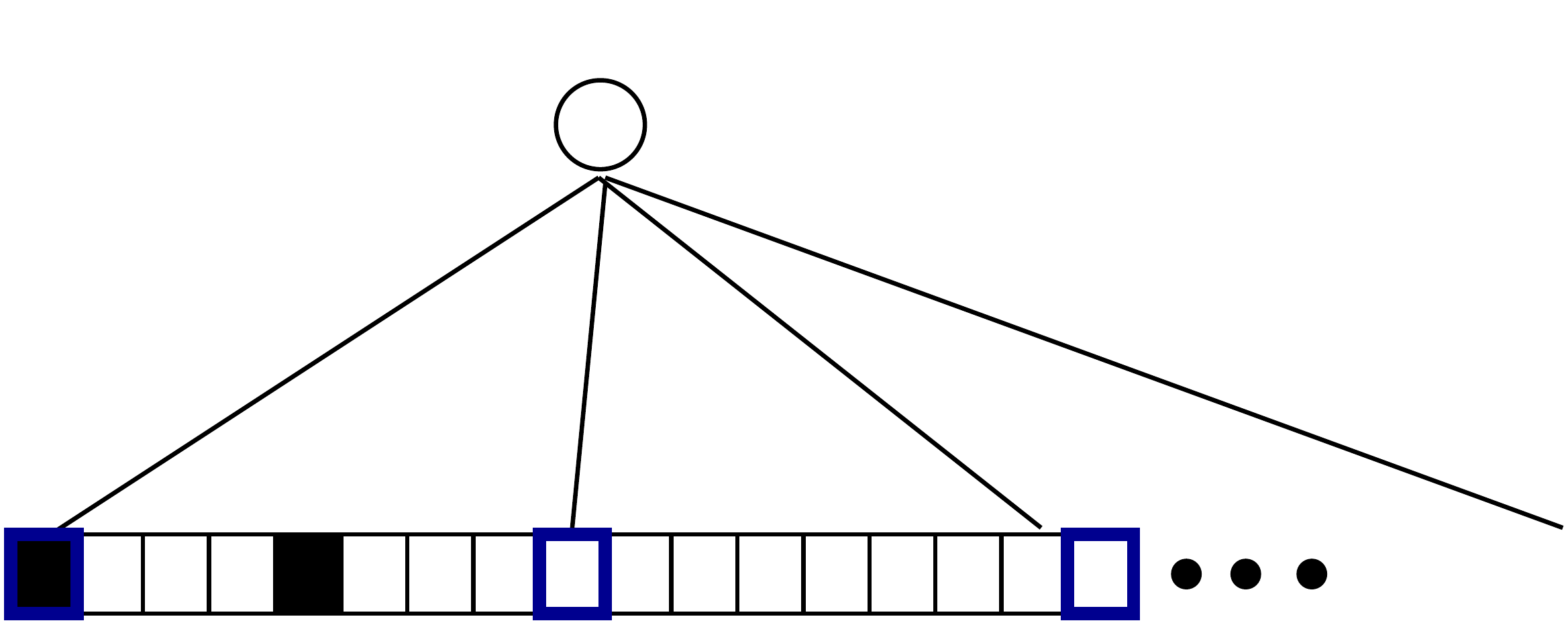} \\
    (a)&(b)&(c)\\
\end{tabular}
\end{center}
\caption{(a) The diagonal of $L:=\bm{x}\bm{y}^\mathrm{T}$ contains evidence for the 
identity transformation. (b) The secondary diagonals contain evidence for shifts.
(c) A hidden unit that pools over one of the diagonals can detect transformations. 
This hidden unit computes a \emph{sum over products}.}
\label{figure:outer}
\end{figure}

It is straightforward to build a content-independent detector if we allow for 
\emph{multiplicative interactions} between the variables.  
In particular, consider the outer product $L:=\bm{x}\bm{y}^\mathrm{T}$ between 
two one-dimensional, binary images, as shown in Figure \ref{figure:outer}. 
Every component $L_{ij}$ of this matrix constitutes evidence for 
exactly one type of transformation (translation, in the example). 
The components $L_{ij}$ act like AND-gates, that can detect \emph{coincidences}. 
Since a component $L_{ij}$ is equal to $1$ only when both corresponding pixels 
are equal to $1$, a hidden unit that pools over multiple components 
(Figure \ref{figure:outer} (c)) is 
much less likely to receive spurious activity that depends on the image content
rather than on the transformation. 
Note that pooling over the components of $L$ amounts to computing 
the \emph{correlation} of the output image with a transformed version of the input image. 
The same is true for real-valued data. 

Based on these observations, a variety of sparse coding models were suggested 
which encode transformations (for example, \cite{olshausenetal, GrimesRao, imtrans}). 
The number of parameters is typically equal to $($the number of hidden variables$)$ $\times$ 
$($the number of input-pixels$)$ $\times$ $($the number of output pixels$)$. 
It is instructional to think of the parameters as populating a $3$-way-``tensor'' $w$ 
with components $w_{ijk}$. 

\begin{figure}
\begin{center}
\begin{tabular}{p{150pt}p{150pt}}
\resizebox{4.5cm}{!}{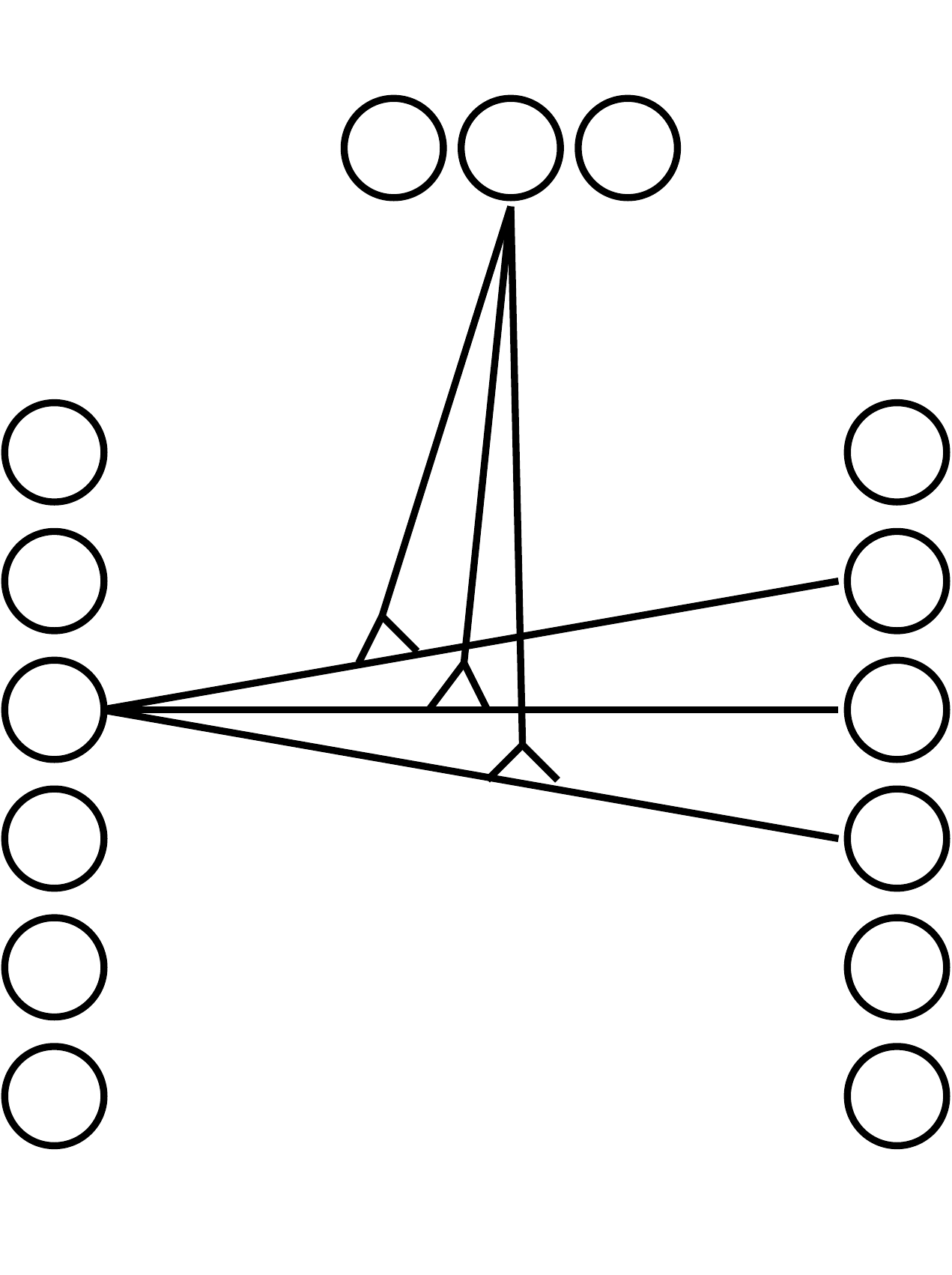}&
\resizebox{5.5cm}{!}{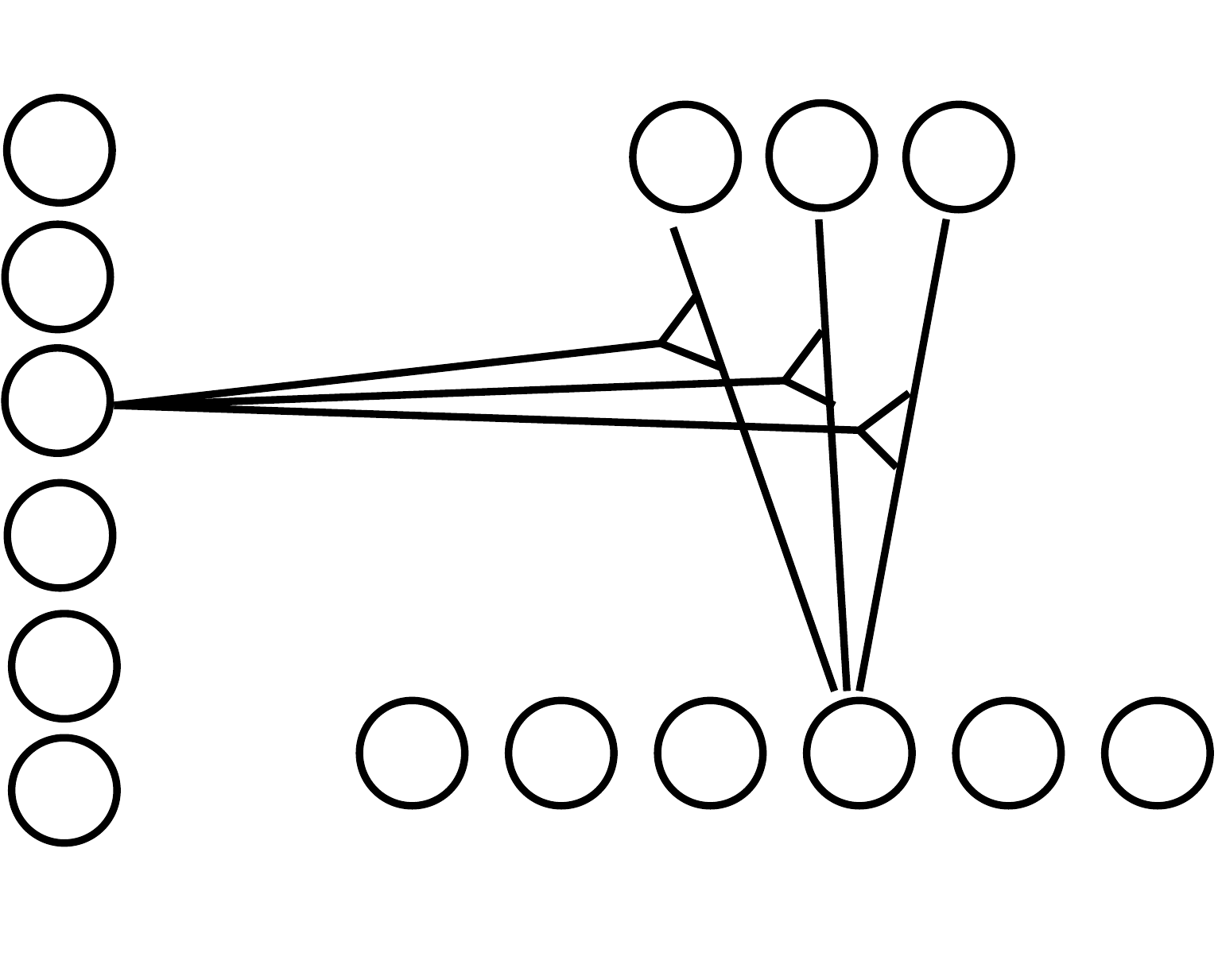}\\
\end{tabular}
\end{center}
\caption{Relating images using multiplicative interactions. Two equivalent views of the same type of model.}
\label{figure:twoviews}
\end{figure}

Figure \ref{figure:twoviews} (left) shows two alternative illustrations of this type of model 
(adapted from \cite{imtrans}). 
Sub-figure (a) shows that each hidden variable can blend in a slice $w_{\cdot\cdot k}$ 
of the parameter tensor. Each slice is a matrix connecting each input pixel to 
each output-pixel. We can think of this matrix as performing linear regression 
in the space of stacked gray-value intensities, known commonly as a ``warp''. 
Thus, the model as a whole can be thought of as defining a \emph{factorial mixture of warps}. 

Alternatively, each input pixel can be thought of as blending in a slice $w_{i\cdot\cdot}$ 
of the parameter tensor. 
Thus, we can think of the model as a standard sparse coding model on the output 
image (Figure \ref{figure:twoviews} (left)), whose parameters are modulated by the 
input image. This turns the model into a \emph{predictive} or \emph{conditional} sparse coding model
\cite{olshausenetal, imtrans}.
In both cases, hidden variables take on the roles of dynamic 
mapping units \cite{MappingsHinton1981, MappingsMalsburg1981} which encode 
the relationship not the content of the images. 
Each unit in the model can gate connections between other variables in the model. We shall 
refer to this type of model as ``gated sparse coding'', or synonymously as ``cross-correlation model''. 

Like in a standard sparse coding model one needs to include biases in practice. 
The set of model parameters thus consists of the three-way parameters $w_{ijk}$, 
as well as of single-node parameters $w_i$, $w_j$ and $w_k$. 
One could also include ``higher-order-biases'' \cite{imtrans}
like $w_{ik}$, which connect two groups of variables, but it is not common to do so. 
Like before, we shall drop all bias terms in what follows in order to 
avoid clutter. Both simple biases and higher-order biases can be implemented 
by adding constant-1 dimensions to data and to hidden variables.

\subsection{Inference}
\label{section:inference}
The graphical model of gated sparse coding models is tri-partite. That of a standard sparse coding 
model is bi-partite.
Inference can be performed in almost the same as in a standard sparse coding 
model, whenever two out of three groups of variables have been observed. 

\begin{figure}
\begin{center}
\begin{tabular}{cc}
    \resizebox{5.0cm}{!}{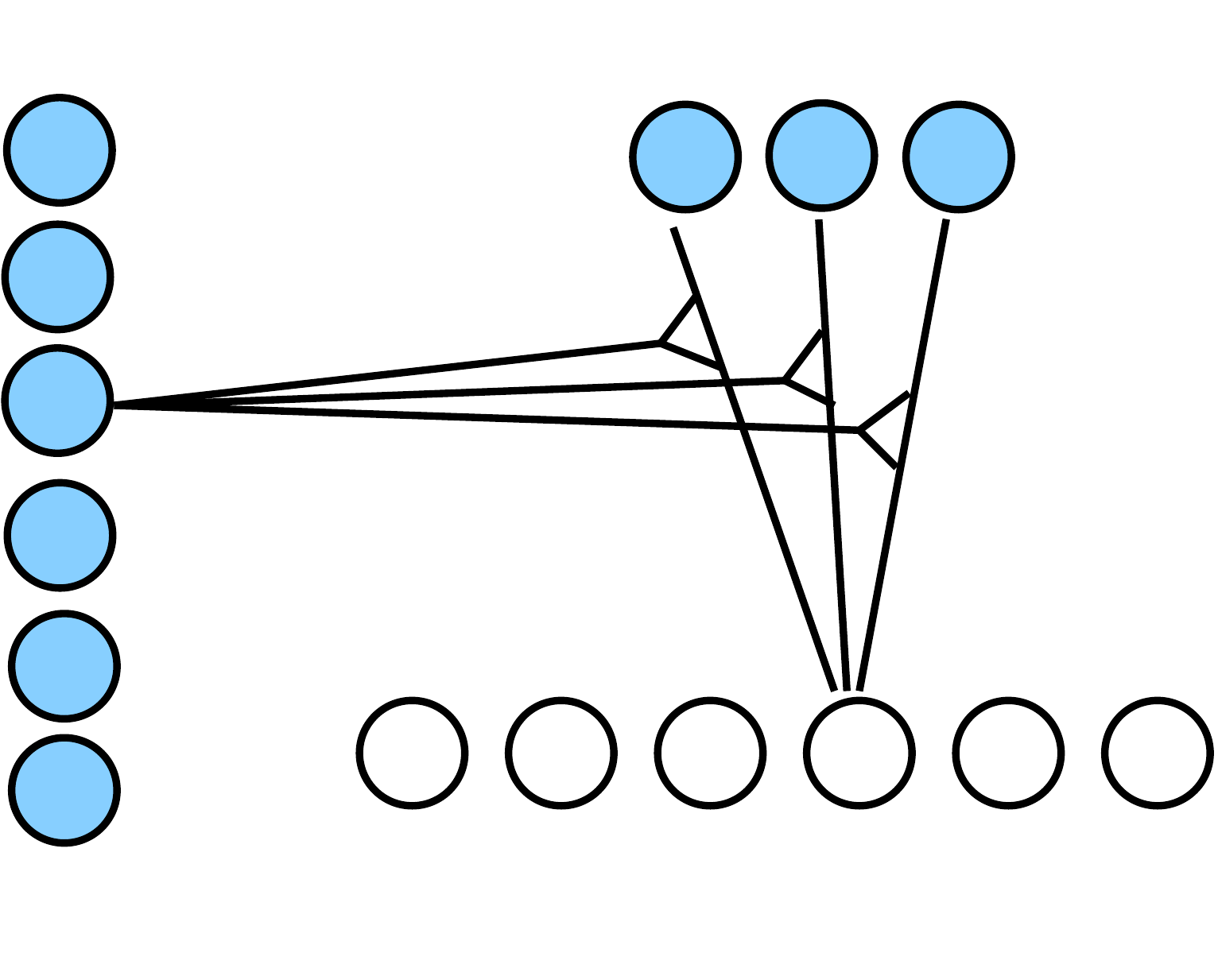} \hspace{50pt} &
    \resizebox{5.0cm}{!}{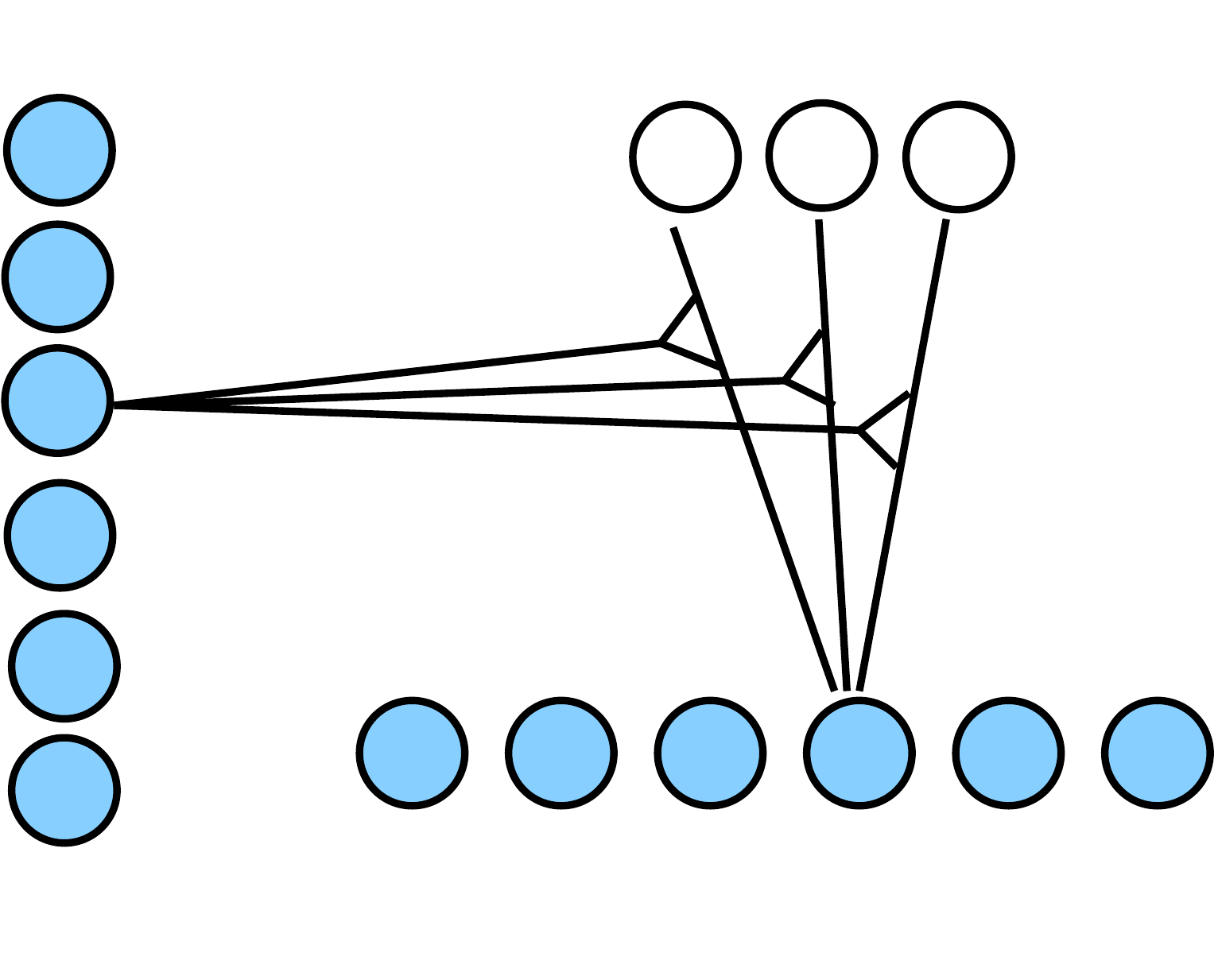} \\
    (a) & (b) \\
\end{tabular}
\end{center}
\caption{Inferring any one group of variables, given the other two, 
is like inference in a standard sparse coding model. Blue shading represents conditioning.}
\label{figure:conditioning}
\end{figure}

Consider, for example, the task of inferring $\bm{z}$, given $\bm{x}$ and $\bm{y}$ 
(see Figure \ref{figure:conditioning} (a)). 
Recall that for a standard sparse coding model, we have: $z_k = \sum_{j} w_{jk} y_j$ 
(up to component-wise non-linearities). 
It is instructional to think of the gated sparse coding model as turning the weights into 
a function of $\bm{x}$. If that function is linear: $w_{jk}(\bm{x})=\sum_i w_{\bm{i}jk} x_i$, 
we get:
\begin{equation}
z_k = \sum_{j} w_{jk} y_j = \sum_j \big( \sum_i w_{ijk} x_i \big) y_j = \sum_{ij} w_{ijk} x_i y_j
\label{equation:inferencez}
\end{equation}
which is exactly of the form discussed in the previous section. 

Eq. \ref{equation:inferencez} shows that 
inference amounts to computing for each output-component $y_j$ a quadratic 
form in $\bm{x}$ and $\bm{z}$  defined by the weight tensor $w_{\cdot j \cdot}$. 
Considering either $\bm{x}$ or $\bm{z}$ as fixed, one can also think of inference as 
a simple linear function like in a standard sparse coding model. This property
is typical of models with bi-linear dependencies \cite{tenenbaum00separating}.
Despite the similarity to a standard sparse coding model, the \emph{meaning} of inference differs from standard sparse coding: 
The meaning of $\bm{z}$, here, is the \emph{transformation} that takes $\bm{x}$ to $\bm{y}$ (or vice versa). 

Inferring $\bm{z}$, given two images $\bm{x}$ and $\bm{y}$ (Figure \ref{figure:conditioning} (b))
yields the analogous expression:
\begin{equation}
y_j = \sum_{k} w_{jk} z_k = \sum_k \big( \sum_i w_{ijk} x_i \big) z_k = \sum_{ik} w_{ijk} x_i z_k, 
\label{equation:inferencey}
\end{equation}
so inference is again a quadratic form.  
The meaning of $\bm{y}$ is now ``$\bm{x}$ transformed according to known transformation $\bm{z}$''. 

For the analysis in Section \ref{section:sharedeigenspaces} it is useful to note that, when 
$\bm{z}$ is given, then $\bm{y}$ is a linear function of $\bm{x}$ (cf. Eq. \ref{equation:inferencey}), so 
it can be written  
\begin{equation}
\bm{y} = L\bm{x}
\label{equation:linearwarp}
\end{equation}
for some matrix $L$, which itself is a function of $\bm{z}$.
Commonly, $\bm{x}$ and $\bm{y}$ represent vectorized images,  
so that the linear function is a \emph{warp}.
Note, that the representation of the linear function is factorial. That is, the hidden variables
make it possible to compose a warp additively from constituting components much like 
a factorial sparse coding model (in contrast to a genuine mixture model) makes it possible to 
compose an image from independent components.  

Like in a standard sparse coding model, it can be useful in some applications to assign a number 
to an input, quantifying \emph{how well it is represented by the model}. 
For this number to be useful, it has to be ``calibrated'', which is typically achieved 
by using a probabilistic model. 
In contrast to a simple sparse coding model, training a probabilistic gated sparse coding model 
can be slightly more complicated, because of the dependencies between $\bm{x}$ 
and $\bm{y}$ conditioned on $\bm{z}$. We discuss this issue in detail in the next section.

\subsection{Learning}
\label{section:learning}
Training data for a gated sparse coding model consists 
of pairs of points $\left(\bm{x}^\alpha, \bm{y}^\alpha \right)$.
Training is similar to standard sparse coding, but there are some important differences. 
In particular, note that the gated model is like a sparse coding model whose input 
is the vectorized outer-product $\bm{x}\bm{y}^\mathrm{T}$ 
(cf. Section \ref{section:modelingrelations}), so that 
standard learning criteria, such as squared error, are obviously not appropriate.

\subsubsection{Predictive training}
One way to train the model is utilizing the view as predictive sparse 
coding (Figure \ref{figure:conditioning} (b)), and to train the model 
conditionally by predicting $\bm{y}$ given $\bm{x}$ \cite{GrimesRao}, \cite{olshausenetal}, \cite{imtrans}.  

Recall that we can think of the inputs $\bm{x}$ as modulating the parameters. 
This modulation is \emph{case-dependent}. 
Learning can therefore be viewed as ``sparse coding with case-dependent weights''. 
The cost that data-case $\left(\bm{x}^\alpha, \bm{y}^\alpha\right)$ contributes is:
\begin{equation}
\sum_j \big( y_j^\alpha - \sum_{ik} w_{ijk} x_i^\alpha z_k^\alpha )^2
\end{equation}
Differentiating with respect to $w_{ijk}$ is the same as in a standard sparse coding model. 
In particular, the model is still \emph{linear} wrt. the parameters. 
Predictive learning is therefore possible with gradient-based optimization 
similar to standard feature learning (cf. Section \ref{section:featurelearning}). 

To avoid iterative inference, it is possible to adapt various sparse coding 
variants, like auto-encoders and RBMs (Section \ref{section:featurelearning}) 
to the conditional case. 
As an example, we obtain a ``gated Boltzmann machine'' (GBM) by changing 
the energy function into the three-way energy \cite{imtrans}:  
\begin{equation}
    E(\bm{x}, \bm{y}, \bm{z}) = \sum_{ijk} w_{ijk} x_i y_j z_k
\end{equation}
and exponentiating and normalizing: 
\begin{equation}
p(\bm{y}, \bm{z}|\bm{x})=\frac{1}{Z(\bm{x})}\exp \big(E(\bm{x}, \bm{y}, \bm{z})\big), \quad
Z(\bm{x})=\sum_{\bm{y},\bm{z}}\exp\big(E(\bm{x}, \bm{y}, \bm{z})\big)
\end{equation}
Note that the normalization is over $\bm{y}$ and $\bm{z}$ only, which is 
consistent with our goal of defining a predictive model. 
It is possible to define a joint model, but this makes training more 
difficult (cf. Section \ref{section:jointtraining}).
Like in a standard RBM, training involves sampling $\bm{z}$ and $\bm{y}$. 
In the relational RBM samples are drawn from the \emph{conditional} distributions 
$p(\bm{y}|\bm{z}, \bm{x})$ and 
$p(\bm{z}|\bm{y}, \bm{x})$.  

\begin{figure}
\begin{center}
\begin{tabular}{c|c}
    \resizebox{6.0cm}{!}{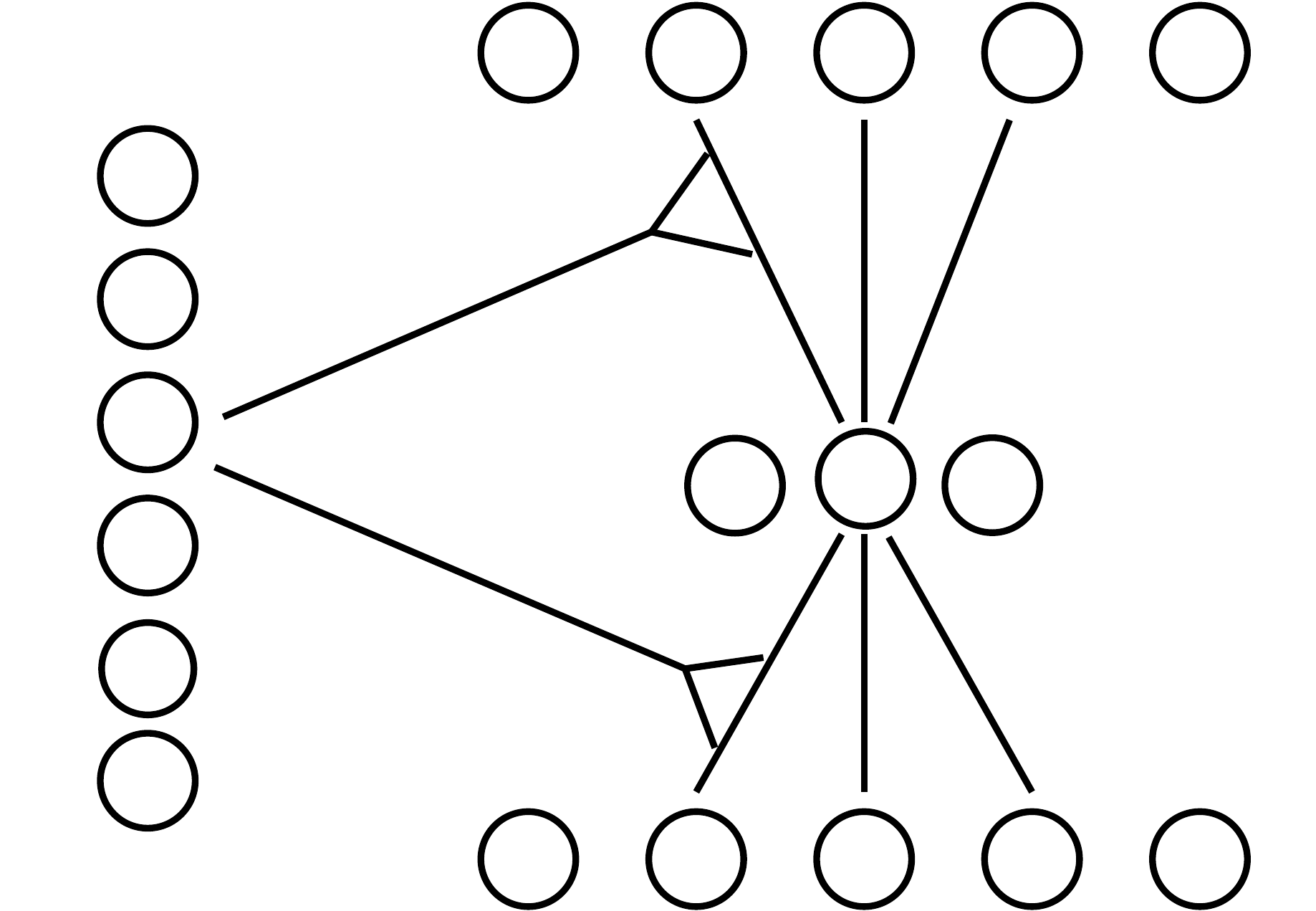} \hspace{20pt}  &
    \resizebox{8.0cm}{!}{\includegraphics[angle=0]{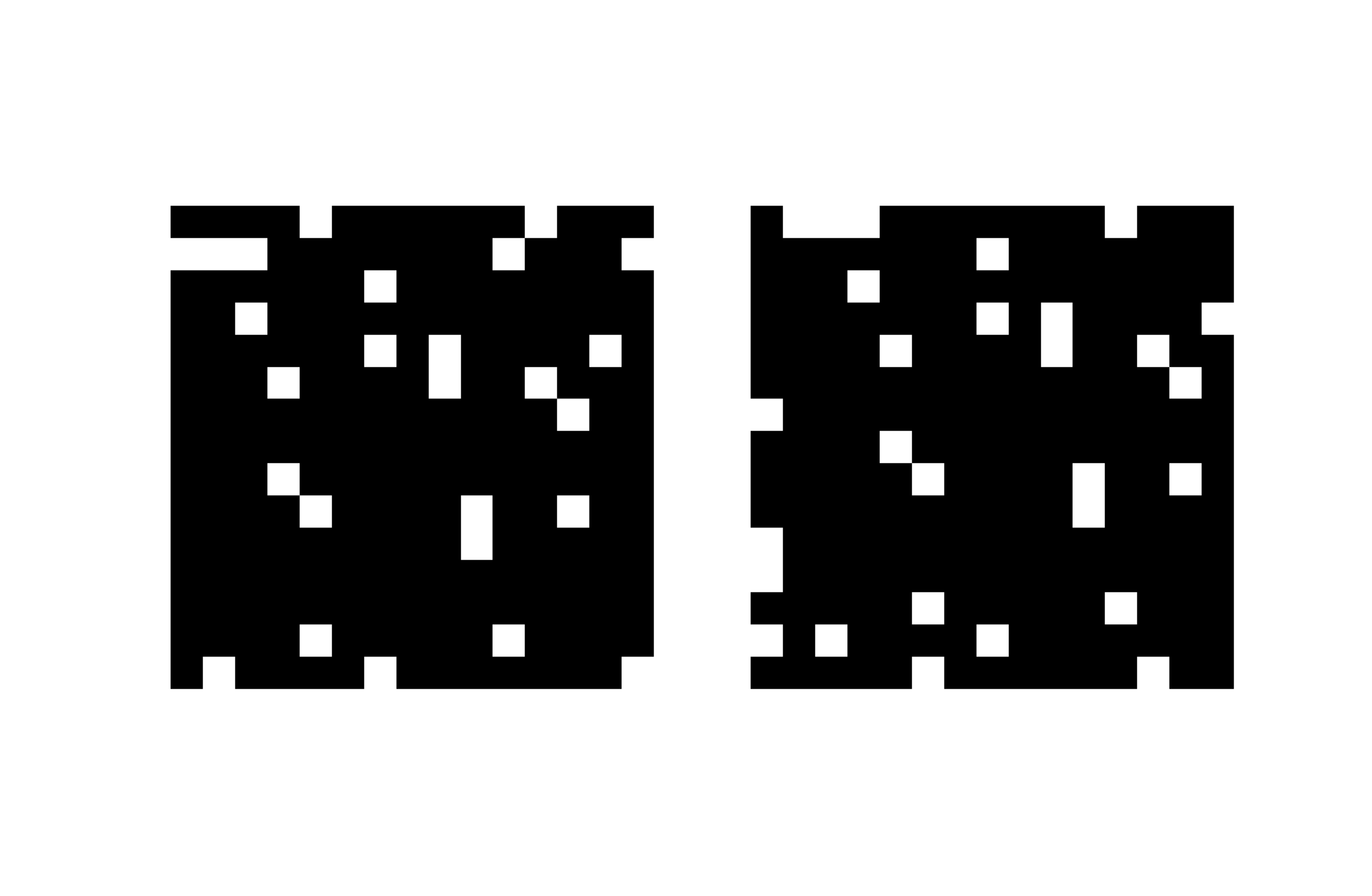}} \\
    (a) & (b) \\
\end{tabular}
\end{center}
\caption{(a) Relational auto-encoder. (b)
Toy data commonly used to test relational models. There is no structure in the images, only in their relationship.
}
\label{figure:rae}
\end{figure}

As another example, we can turn an auto-encoder into a \emph{relational auto-encoder}, 
by defining the encoder and decoder parameters $A$ and $W$ as linear functions of $\bm{x}$ 
(\cite{memisevic08}, \cite{higherordergradientbased}). 
Learning is then essentially the same as in a standard auto-encoder 
modeling $\bm{y}$. In particular, the model is still a directed acyclic graph, 
so one can use simple back-propagation to train the model. 
See Figure \ref{figure:rae} for an illustration.

\subsubsection{Symmetric training}
\label{section:jointtraining}
In probabilistic terms, predictive training amounts to modeling 
the \emph{conditional distribution} 
$p(\bm{y}|\bm{x})=\int_{\bm{z}} p(\bm{y},\bm{z}|\bm{x}) \; d\bm{z}$. 
\cite{jointImtrans} show how modeling instead the \emph{joint distribution}
can make it possible to perform image matching, by allowing us to quantify 
how compatible any two images are under to the trained model.

Formally, modeling the joint amounts simply to changing the normalization constant 
of the three-way RBM to 
$Z=\sum_{\bm{x}, \bm{y},\bm{z}}\exp\big(E(\bm{x}, \bm{y}, \bm{z})\big)$ (cf. previous section). 
Learning is more complicated, however, because the simplifying view of 
case-based modulation no longer holds. \cite{jointImtrans} suggest 
using three-way Gibbs sampling to train the model.

As an alternative to modeling a joint probability distribution, \cite{higherordergradientbased} show how 
one can instead use a relational auto-encoder trained symmetrically 
on the \emph{sum of the two predictive objectives}
\begin{equation}
\sum_j \big( y_j^\alpha - \sum_{ik} w_{ijk} x_i^\alpha z_k^\alpha )^2 + 
\sum_i \big( x_i^\alpha - \sum_{jk} w_{ijk} y_j^\alpha z_k^\alpha )^2 
\end{equation}
This forces parameters to be able to transform in both directions, and it can give 
performance similar to symmetrically trained, fully probabilistic models \cite{higherordergradientbased}. 
Like an auto-encoder, this model can be trained with gradient based optimization. 

\subsubsection{Learning higher-order within-image structure}
Another reason for learning the joint distribution is that it allows us to 
model higher-order \emph{within-image structure} (for example, \cite{Karklin-Lewicki-06-NIPS, RAN10, HyvarinenISA}). 

\cite{RAN10} apply a GBM to the task of modeling second-order within-image features, 
that is, features that encode pair-wise products of pixel intensities. 
They show that this can be achieved by optimizing the joint GBM distribution 
and using \emph{the same} image as input $\bm{x}$ and as output $\bm{y}$. 
In contrast to \cite{jointImtrans}, 
\cite{RAN10} suggest hybrid Monte Carlo to train the joint.

One can also combine higher-order models with standard sparse coding models,
by using some hidden units to model higher-order structure and some to 
learn linear codes \cite{mcrbm, higherordergradientbased}.

\subsubsection{Toy example: Motion extraction and analogy making}
\begin{figure}
\begin{center}
\begin{tabular}{cc}
     \scalebox{0.112}[0.112]{\includegraphics[angle=0]{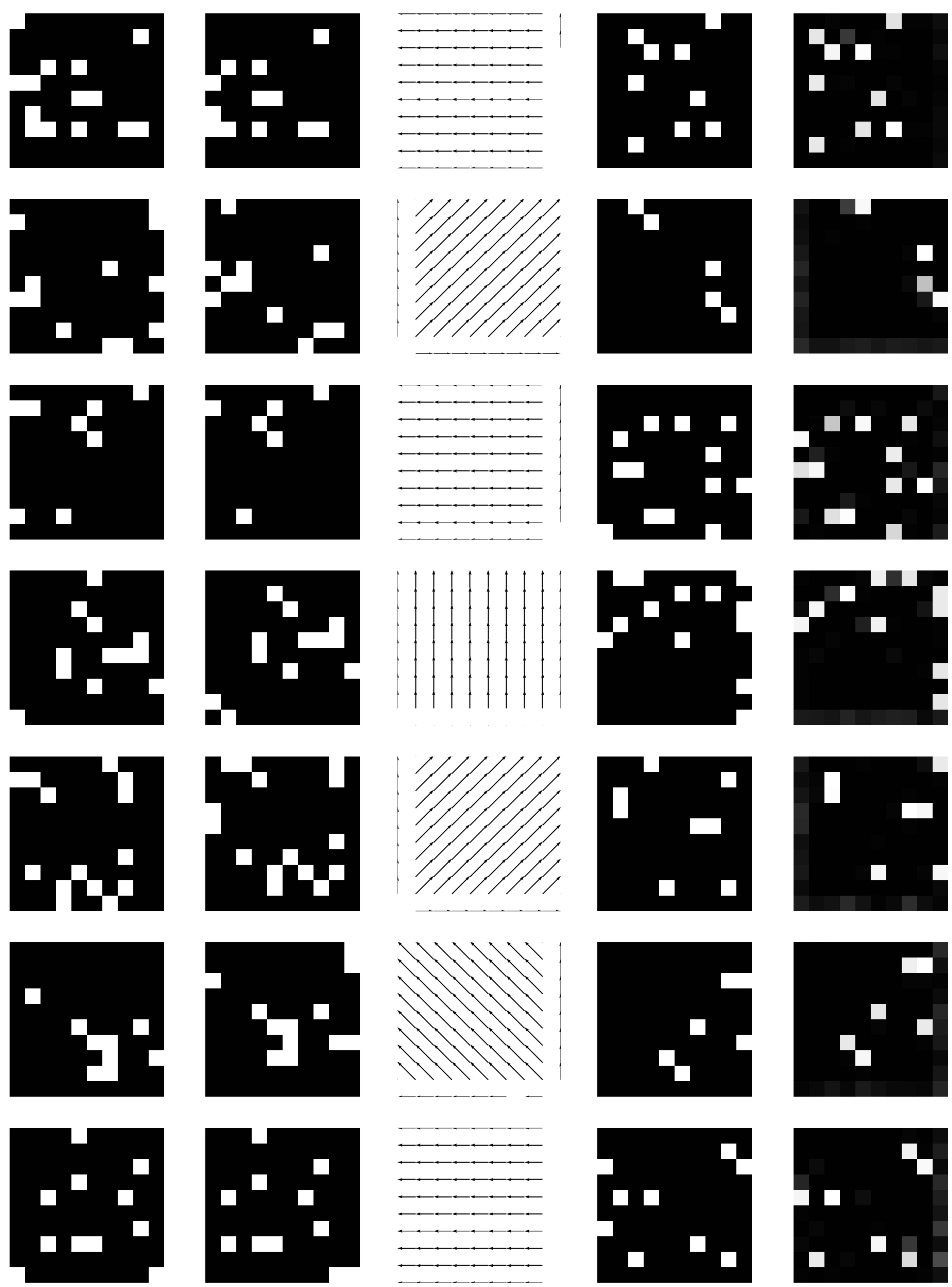}} \hspace{10pt} &
     \scalebox{0.114}[0.114]{\includegraphics[angle=0]{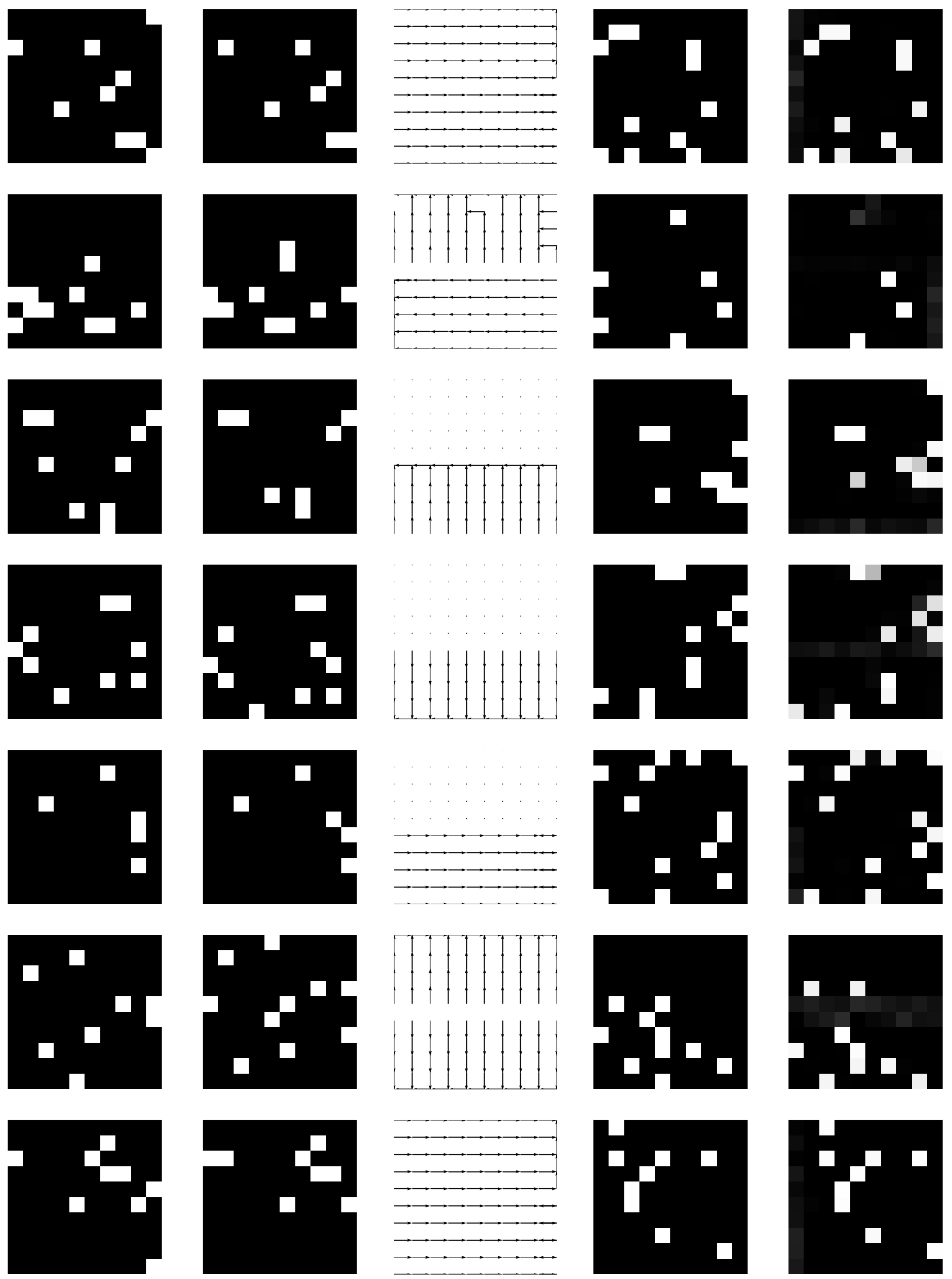}}\\
    (a) & (b) \\
\end{tabular}
\end{center}
\caption{Inferring motion direction from test data. (a) Coherent motion across the whole image.
(b) ``Factorial motion'' that is independent in different image regions.
In both plots, the meaning of the five columns is as follows (left-to-right):
Random test images $\bm{x}$, random test images $\bm{y}$, inferred flow-field, 
new test-image $\hat{\bm{x}}$, inferred output $\hat{\bm{y}}$. 
}
\label{figure:flowexamples}
\end{figure}

Figure \ref{figure:flowexamples} (a) shows a toy example of a gated Boltzmann machine 
applied to translations. 
The model was trained on images showing iid random dots where the output image $\bm{y}$ is a 
copy of the input image $\bm{x}$ shifted in a random direction. The center column in both plots in 
Figure \ref{figure:flowexamples} visualizes the inferred transformation as a 
vector field. The vector-field was produced by (i) inferring the transformation 
given the image pair (Eq. \ref{equation:inferencez}), (ii) computing the 
transformation from the inferred hiddens, and (iii) finding for each 
input-pixel the output-position it is most strongly connected to \cite{imtrans}. 
The two right-most columns in both plots show how the inferred transformation 
can be applied to new images \emph{by analogy}, that is, by computing the output-image 
given a new input image and the inferred transformation (Eq. \ref{equation:inferencey}). 
Figure \ref{figure:flowexamples} (b) shows an example, where the transformations 
are split-screen translations, that is, translations which are independent in the 
top half vs. the bottom half of 
the image. This illustrates how the model has to decompose transformations into factorial 
constituting transformations.

\section{Factorization and energy models}
\label{section:factorization}
In the following, we discuss the close relationship between gated sparse coding models and energy models. 
For this end, we first describe how parameter factorization makes it possible to pre-process input images
and thereby reduce the number of parameters.

\subsection{Factorizing the gating parameters}
The number of gating parameters is roughly cubic in 
the number of pixels, if we assume that the number
of constituting transformations is about the same as the number of pixels.
It can easily be more for highly over-complete hiddens.
\cite{factoredGBM} suggest reducing that number by 
factorizing the parameter tensor $W$ into three matrices, such that each component 
$w_{ijk}$ is given by the ``three-way inner product''
\begin{equation}
\label{equation:factorization}
w_{ijk} = \sum_{ijk} \sum_{f=1}^F w^x_{if}w^y_{jf}w^z_{kf} 
\end{equation}
Here, $F$ is a number of hidden ``factors'', which, like the number $K$ of hidden units, has to be chosen 
by hand or by cross-validation. 
The matrices $w^x$, $w^y$ and $w^z$ are $I \times F$, $J \times F$ and $K \times F$, respectively.

\begin{figure}
\begin{tabular}{cc}
\resizebox{7.5cm}{!}{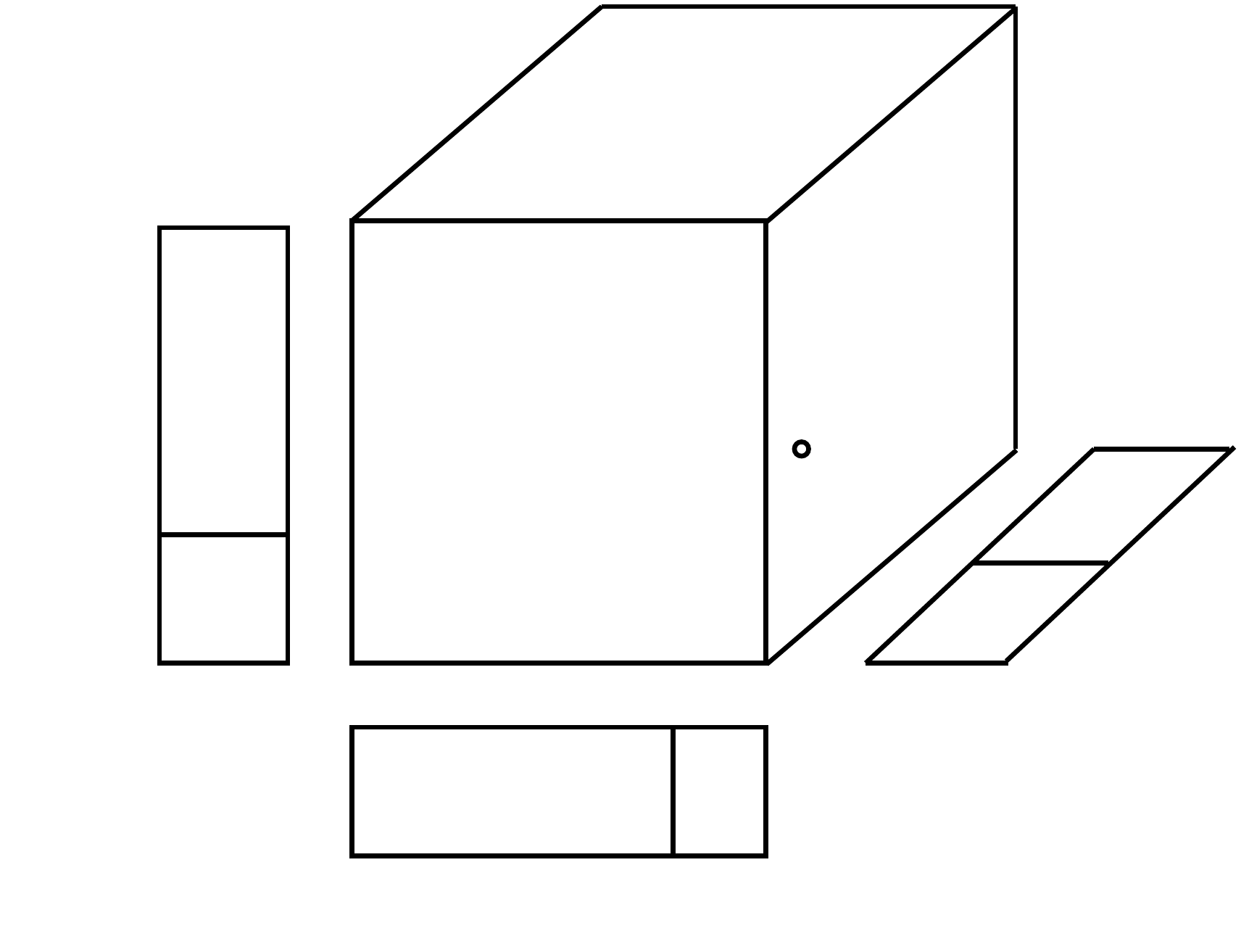} \hspace{10pt} &
\resizebox{6.0cm}{!}{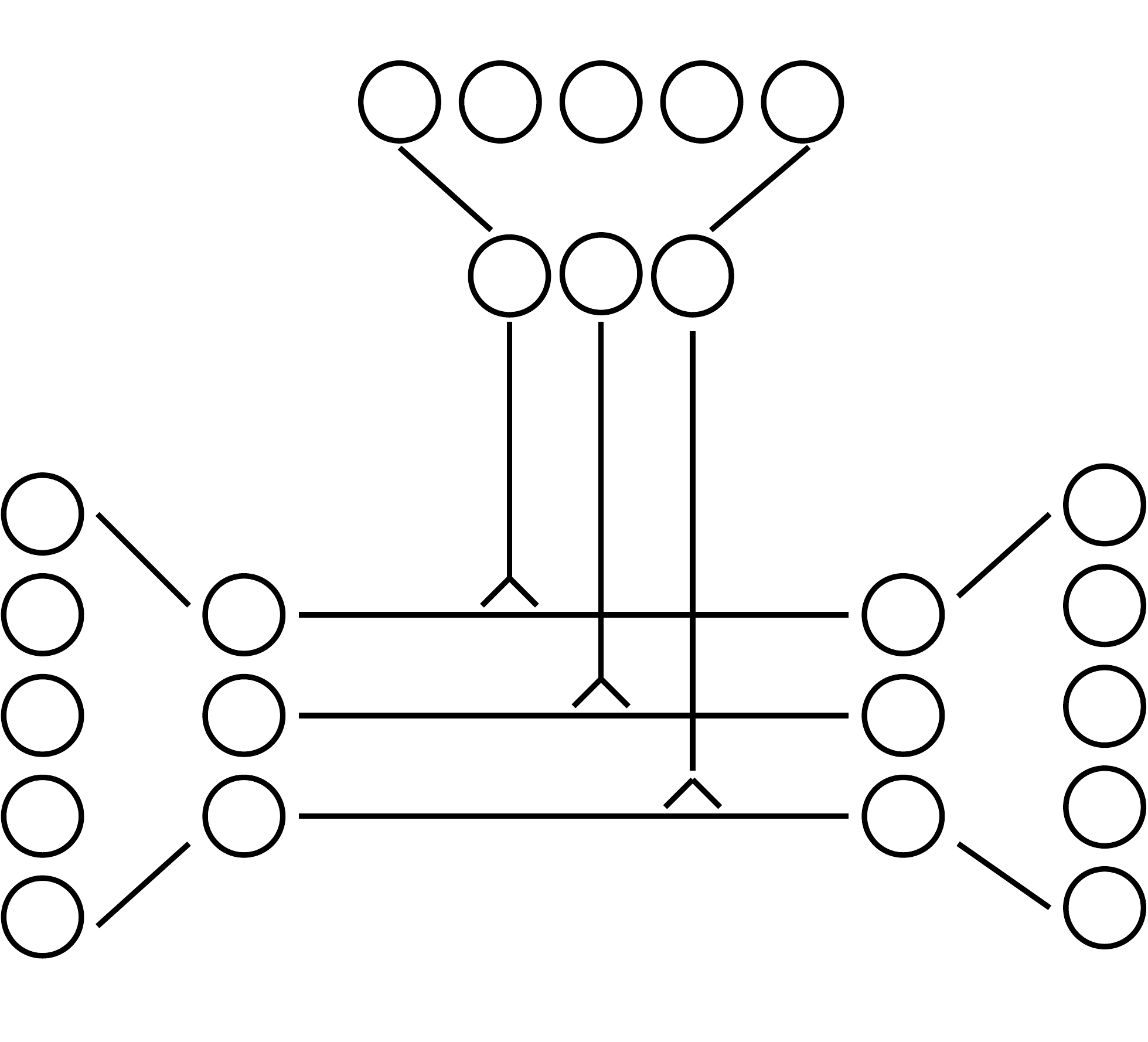}\\
(a)&(b)\\
\end{tabular}
\caption{(a) Factorizing the parameter tensor. (b) Interpreting factorization as \emph{filter matching}.}
\label{figure:factorization}
\end{figure}

An illustration of this factorization is given in Figure \ref{figure:factorization} (a).
It is interesting to note that, under this factorization, the activity of output-variable $y_j$,
by using the distributive law, can by written: 
\begin{equation}
    y_j = \sum_{ik} w_{ijk} x_i z_k=\sum_{ik} (\sum_f w^x_{if}w^y_{jf}w^z_{kf}) x_iz_k =\sum_f w^y_{jf} (\sum_i w^x_{if}x_i) (\sum_k w^z_{kf}z_k)
\label{equation:factorizationisfiltermatching_prediction}
\end{equation}
Similarly, for $z_k$ we have  
\begin{equation}
    z_k = \sum_{ij} w_{ijk} x_i y_j=\sum_{ij} (\sum_f w^x_{if}w^y_{jf}w^z_{kf}) x_iy_j =\sum_f w^z_{kf} (\sum_i w^x_{if}x_i) (\sum_j w^y_{jf}y_j)
\label{equation:factorizationisfiltermatching_inference}
\end{equation}
One can obtain a similar expression for the energy in a gated Boltzmann machine. 
Eq. \ref{equation:factorizationisfiltermatching_inference} shows that factorization can 
be viewed as \emph{filter matching}: For inference, each group 
of variables $\bm{x}$, $\bm{y}$ and $\bm{z}$ are projected onto linear basis functions 
which are subsequently multiplied, as illustrated in Figure \ref{figure:factorization} (b).  

It is important to note that the way factorization reduces parameters is \emph{not} by 
projecting data onto a lower-dimensional space before computing the multiplicative interactions -- 
a claim that can be found frequently in the literature. 
In fact, frequently, $F$ is chosen to be \emph{larger} than $I$ and/or $J$. The way that 
factorization reduces the number of parameters is by restricting \emph{three-way connectivity}.
Learning then amounts to finding basis functions that can deal with this restriction optimally. 
Using the factorization in Eq. \ref{equation:factorization} amounts to allowing each 
factor to engage only in \emph{a single} multiplicative interaction. 

All gated sparse coding models can be subjected to this factorization. 
Training is similar to training an unfactored model by using the chain rule and differentiating 
Eq. \ref{equation:factorization}.  
An example of a factored gated auto-encoder is described in \cite{higherordergradientbased}. 
Virtually all factored models that were introduced use the restriction of single multiplicative 
interactions (Eq. \ref{equation:factorization}). 
An open research question is to what degree a less restrictive connectivity -- 
equivalently, using a non-diagonal core-tensor in the factorization -- would be advantageous. 

\begin{figure}[t!]
\begin{center}
    \begin{tabular}{cc}
        \scalebox{0.54}[0.54]{\includegraphics[angle=0]{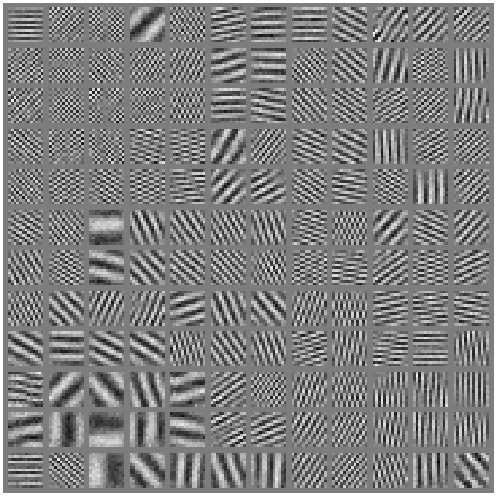}}&
        \scalebox{0.54}[0.54]{\includegraphics[angle=0]{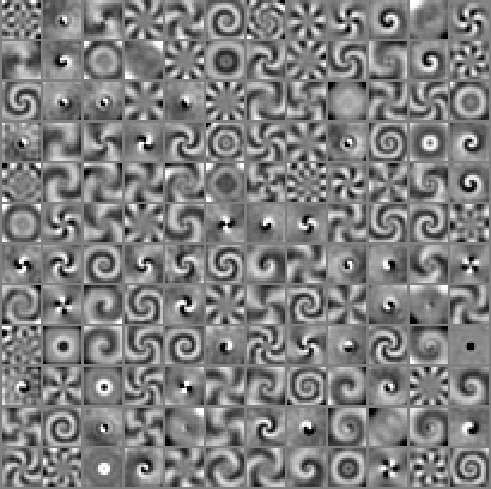}}\\
        \scalebox{0.165}[0.165]{\includegraphics[angle=0]{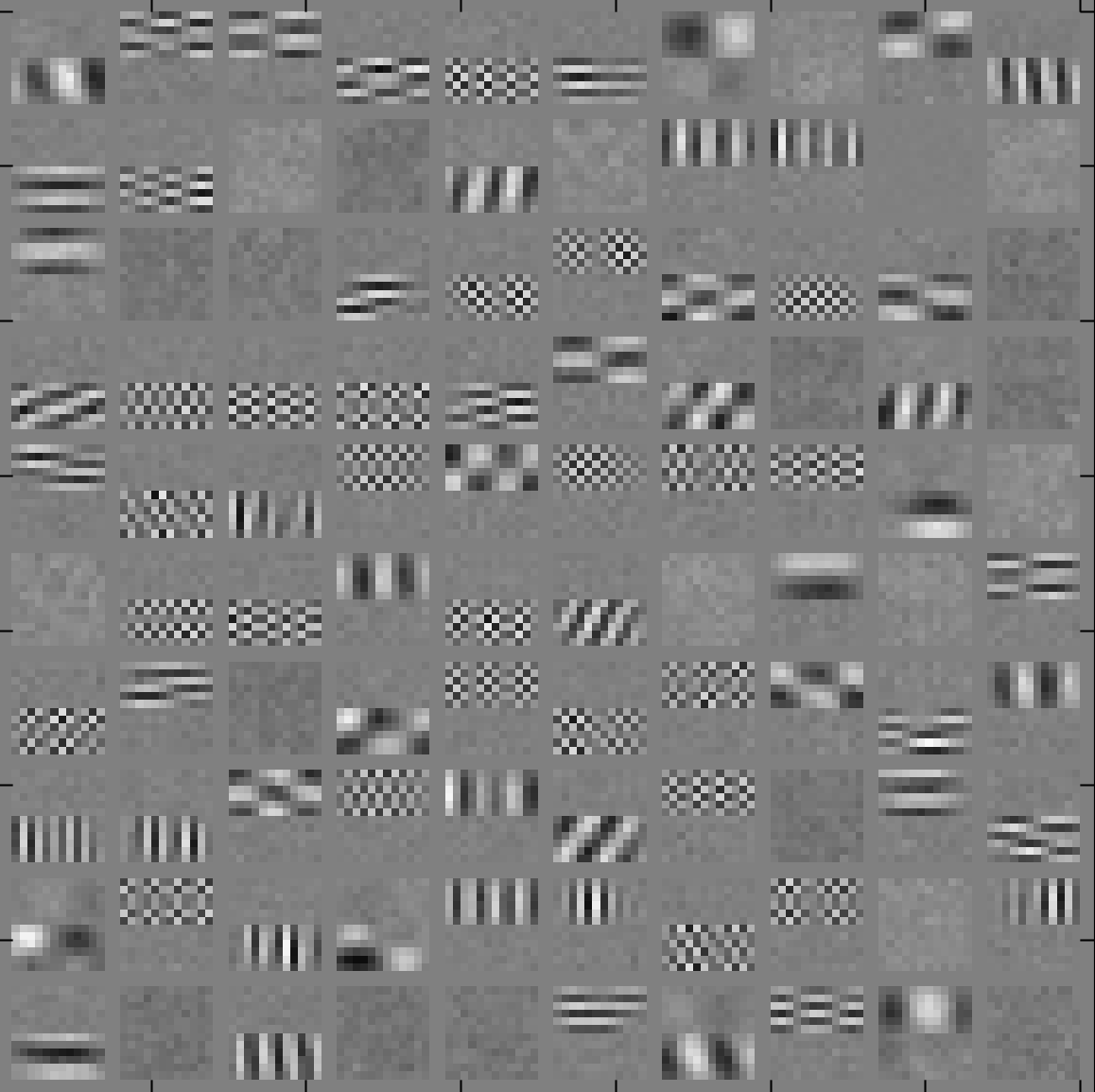}}&
        \scalebox{0.49}[0.49]{\includegraphics[angle=0]{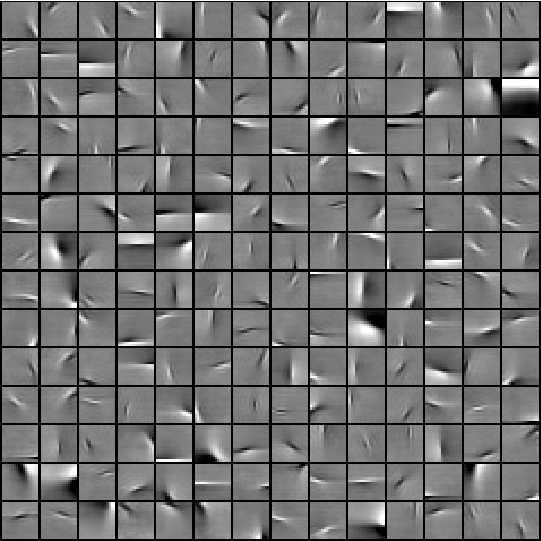}}\\
    \end{tabular}
\end{center}
\caption{Input filters learned from various types of transformation. Top-left: Translation, Top-right: Rotation, Bottom-left: split-screen translation, Bottom-right: Natural videos. See figure \ref{figure:scrollfiltersy} on the next page for corresponding output filters.}
\label{figure:scrollfiltersx}
\end{figure}

\begin{figure}[t!]
\begin{center}
    \begin{tabular}{cc}
        \scalebox{0.54}[0.54]{\includegraphics[angle=0]{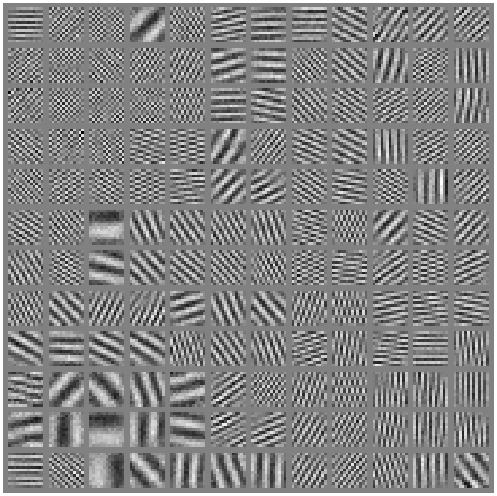}}&
        \scalebox{0.54}[0.54]{\includegraphics[angle=0]{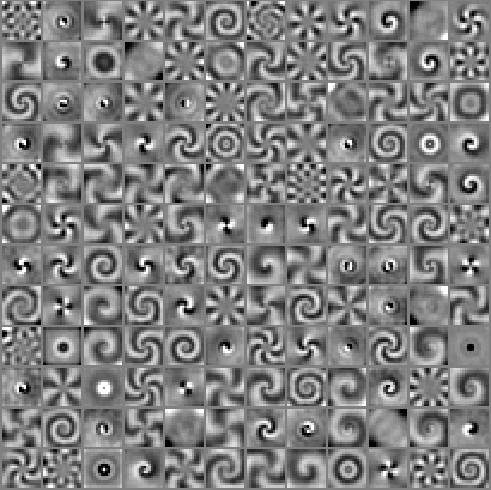}}\\
        \scalebox{0.165}[0.165]{\includegraphics[angle=0]{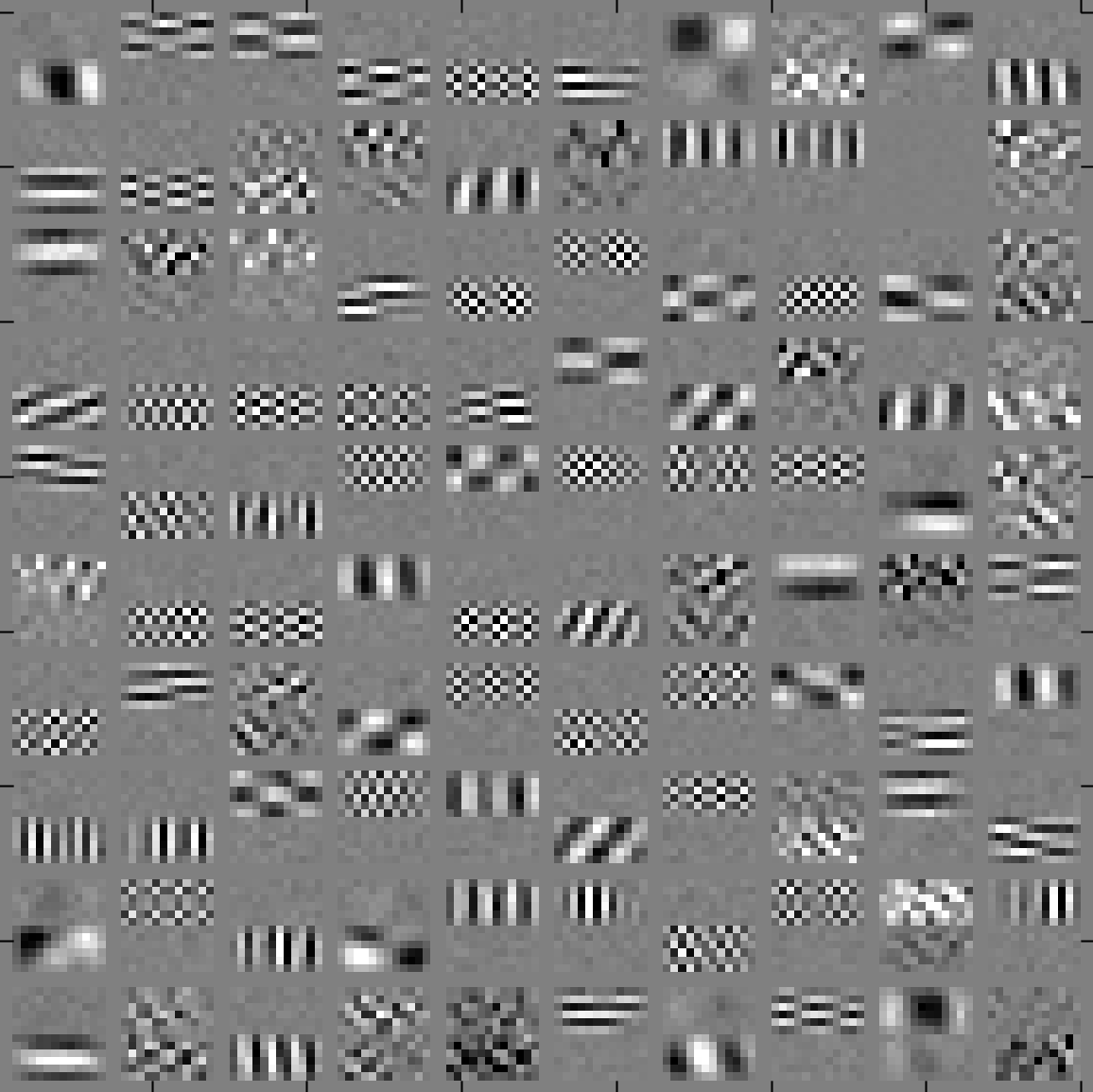}}&
        \scalebox{0.49}[0.49]{\includegraphics[angle=0]{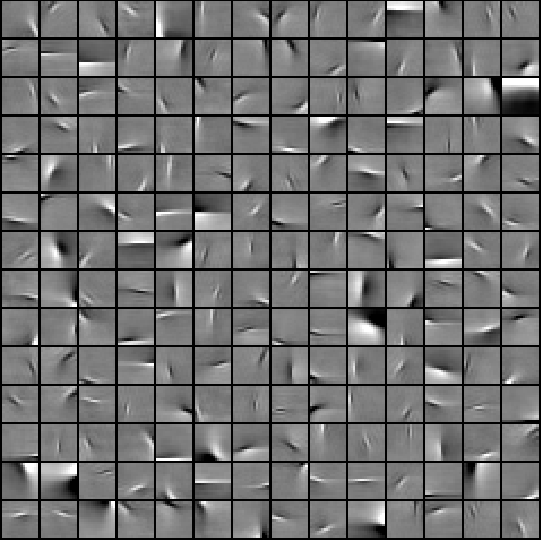}}\\
    \end{tabular}
\end{center}
\caption{Output filters learned from various types of transformation. Top-left: Translation, Top-right: Rotation, Bottom-left: split-screen translation, Bottom-right: Natural videos. See figure \ref{figure:scrollfiltersx} on the previous page for corresponding input filters.}
\label{figure:scrollfiltersy}
\end{figure}

\cite{factoredGBM} show empirically how training factored model leads to filter-pairs that optimally 
represent transformation classes, such as Fourier-components for translations and a polar 
variant of Fourier-components for rotations. 
Figures \ref{figure:scrollfiltersx} and \ref{figure:scrollfiltersy} show 
examples of filters learned from translations, affine transformations, 
split-screen translations, which are independent in the top and bottom half of the image, 
and natural video. 
For training the filters in the top rows and on the bottom right, we used data-sets 
described in \cite{factoredGBM} and \cite{higherordergradientbased} and the model described 
in \cite{higherordergradientbased}. 
The filters resemble receptive fields found in various cells in visual cortex \cite{GallantBraunVanEssen93}. 
To obtain split-screen filters (bottom left) we generated a data-set of split-screen 
translations and trained the model described in \cite{factoredGBM}. 
In Section \ref{section:sharedeigenspaces}, we provide an analysis that sheds some light 
onto why the filters take on this form.

\subsection{Energy models}
\label{section:energymodels}
Energy models \cite{adelson1985spatiotemporal, ODF} are an alternative approach to 
modeling image motion and disparities, and they have been deployed monocularly, too. 
A main application of energy models is the detection of small translational motion 
in image pairs. This makes them suitable as biologically 
plausible mechanisms of both local motion estimation and binocular disparity estimation. 
Energy models detect motion by projecting two images onto two phase-shifted 
Gabor functions each (for a total of four basis function responses).  
The two responses \emph{across} the images are added and squared. 
The sum of these two squared, spatio-temporal responses then yields the response of the energy model. 

The rationale behind the energy model is that, since each within-image Gabor filter pair 
can be thought of as a localized spatio-temporal Fourier component, the sum of the squared 
components yields an estimate of spectral energy, which is not dependent of the \emph{phase} --
and thus to a large degree not dependent on the \emph{content} -- of the input images. 
The two filters within each image need to be sine/cosine pairs, which is commonly referred to 
as being ``in quadrature''.  

A detector of local shift can be built by using a set of energy models tuned to different 
frequencies. To turn the set of energy responses into an estimate of local translation, one can,  
for example, pick the model with the strongest response \cite{sangerStereoGabor, QianStereoAndMotion}, 
or use pooling to get a more stable estimate \cite{FleetBinocular}.

\begin{figure}
\begin{center}
    \resizebox{5.0cm}{!}{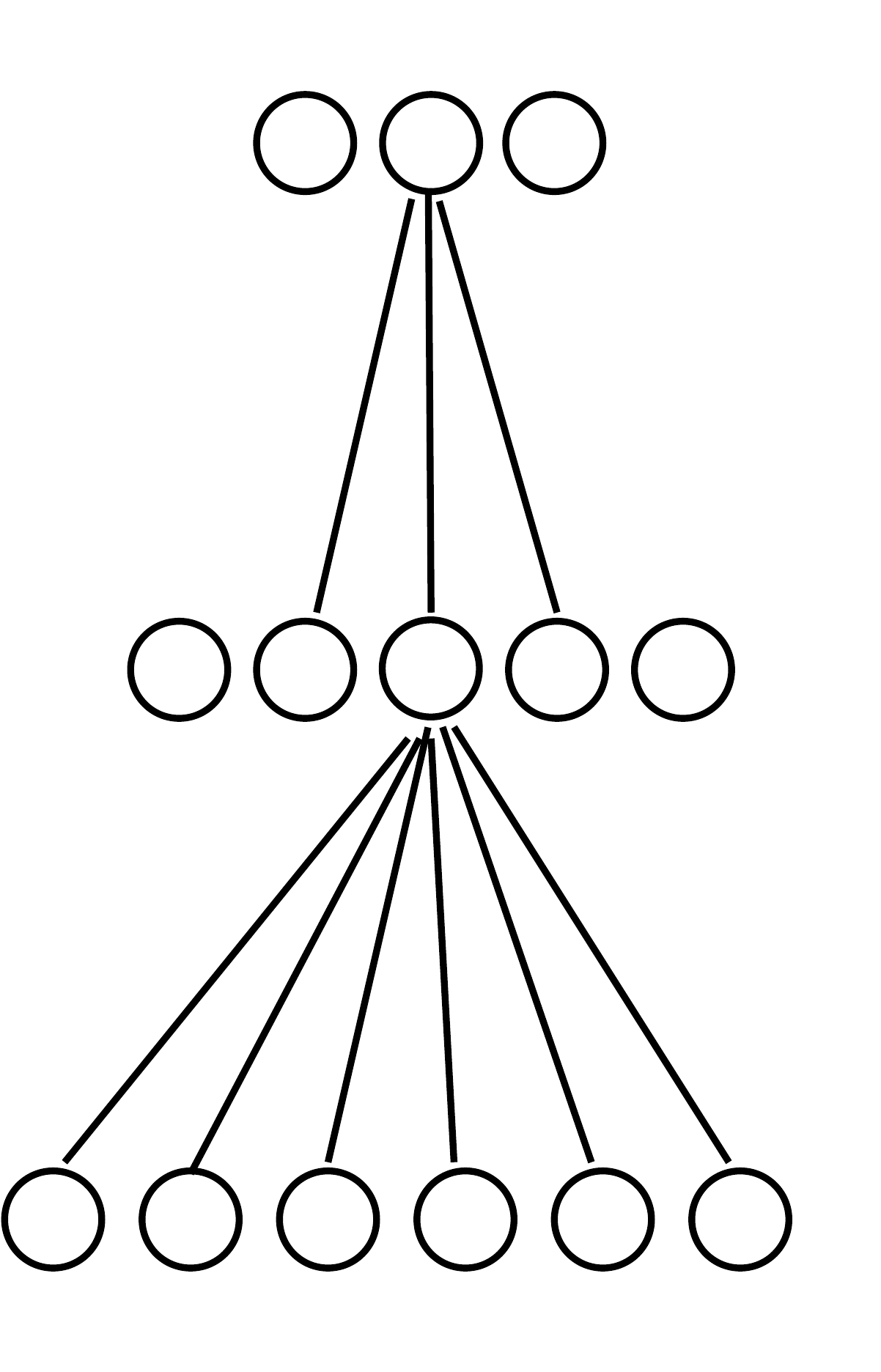}
\end{center}
\caption{Illustration of Independent Subspace Analysis applied to an image pair $(\bm{x}, \bm{y})$.}
\label{figure:isa}
\end{figure}

\cite{Kohonen1995, HyvarinenISA} suggest \emph{learning} energy-like models from data by 
extending a sparse coding model with an elementwise squaring operation, followed by a 
linear pooling layer. 
In contrast to the original energy model, one may use more than exactly two filters to 
pool over, and pooling weights may be learned along with basis functions, instead of 
being fixed to be $1$. 
Figure \ref{figure:isa} shows an illustration of this type of model applied to an image pair. 
As the figure shows, this type of model can be viewed as a two-layer network, 
with a hidden layer that uses an elementwise squaring nonlinearity. 

For learning, \cite{HyvarinenISA} suggest adopting ICA 
by forcing the responses of latent variables (which are 
now sums of squared basis function responses) 
to be sparse, while keeping the filters orthogonal to avoid degenerate solutions, 
just like when training a standard ICA model (cf. Section \ref{section:featurelearning}). 
This approach is known as ``Independent Subspace Analysis'' (ISA). 
We shall refer to the hidden layer nodes as ``factors'' in 
analogy to the hidden layer of a factored GBM. 
Both ISA and factored gated Boltzmann machines were shown to yield state-of-the-art 
performance in various motion recognition tasks \cite{201106-cvpr-le, tayloreccv10}.

\subsection{Relationship between gated sparse coding and energy models}
\label{section:relationsparsecodingenergymodels}
Learning energy models, such as ISA, \emph{on the concatenation of two inputs $\bm{x}$ and $\bm{y}$} 
is closely related to learning gated sparse coding models.  
Let $w_{\cdot f}^x$ ($w_{\cdot f}^y$) denote the set of weights connecting part $\bm{x}$ ($\bm{y}$)
of the concatenated input with factor $f$ (cf. Figure \ref{figure:isa}). 
The activity of hidden unit $z_k$ in an energy model is given by 
\begin{eqnarray}
\label{equation:energycrosscorrelation}
z_k&=&\sum_f w_{kf}^z \big({w^x_{\cdot f}}^\mathrm{T}\bm{x}+{w^y_{\cdot f}}^\mathrm{T}\bm{y}\big)^2\\
\label{equation:energycrosscorrelation2}
&=&\sum_f w^z_{kf} \big( 2 ({w^x_{\cdot f}}^\mathrm{T} \bm{x}) ({w^y_{\cdot f}}^\mathrm{T}\bm{y}) + ({w^x_{\cdot f}}^\mathrm{T} \bm{x})^2 + ({w^y_{\cdot f}}^\mathrm{T} \bm{y})^2\big)
\end{eqnarray}
%
Up to the quadratic terms in Eq. \ref{equation:energycrosscorrelation2}, hidden unit
activities are the same as in a gated sparse coding model. As we shall discuss in 
detail below, the quadratic terms do not have a significant effect on the meaning 
of the hidden units. They can therefore also be thought of as a way to implement mapping 
units that encode relations.

\subsection{Implementing gated sparse coding models}
Over the years, a variety of tricks and recipes have emerged, which can simplify, 
stabilize, or speed up, learning in the presence of multiplicative interactions.  
One approach, that is used by practically everyone in the field, is to normalize 
output filter matrices $w^x$ and $w^y$ 
during learning, such that all filter $w^x_{\cdot f}$ and $w^y_{\cdot f}$ grow slowly
and maintain roughly the same length as learning progresses. 
A common way to achieve this is to maintain a running average of the \emph{average norm} 
of the filters during learning and to re-normalize each filter to have this norm after 
every learning update. 
Furthermore, it is common to connect top-level hidden units \emph{locally} to the factors,
rather than using full connectivity. The theoretical discussion in the next section 
provides some intuition into why local connectivity helps speed up learning. 
A slightly more complicated approach is to let all hidden units populate a 
virtual ``grid'' in a low-dimensional space (for example, 2-D) and to connect 
hidden units to factors, such that neighboring hidden units are connected to the same 
or to overlapping sets of factors. 
The approach has been popular mainly in the context of learning energy models 
(for example, \cite{Welling02, HyvarinenHI00}). 
Finally, it is common to train the models using image patches that are DC centered and 
contrast normalized, and usually also whitened.

\section{Relational codes and simultaneous eigenspaces}
\label{section:sharedeigenspaces}
We now show that hidden variables learn to detect subspace-rotations when they are trained on 
transformed image pairs.  
In Section \ref{section:inference} (Eq. \ref{equation:linearwarp}) we showed that
transformation codes $\bm{z}$ can represent linear transformations, $L$, that is 
$\bm{y}=L\bm{x}$.
We shall restrict our attention in the following to transformations, $L$, that are orthogonal,
that is, $L^\mathrm{T}L=LL^\mathrm{T}=I$, where $I$ is the 
identity matrix. In other words, $L^{-1}=L^\mathrm{T}$. 
Linear transformations in ``pixel-space'' are also known as \emph{warp}.
Note that practically all relevant spatial transformations, like translation, rotation or 
local shifts, can be expressed approximately as an orthogonal 
warp, because orthogonal transformations subsume, in particular, all \emph{permutations} (``shuffling pixels''). 

\begin{figure*}
    \begin{center}
    \begin{tabular}{cc}
        \resizebox{8.3cm}{!}{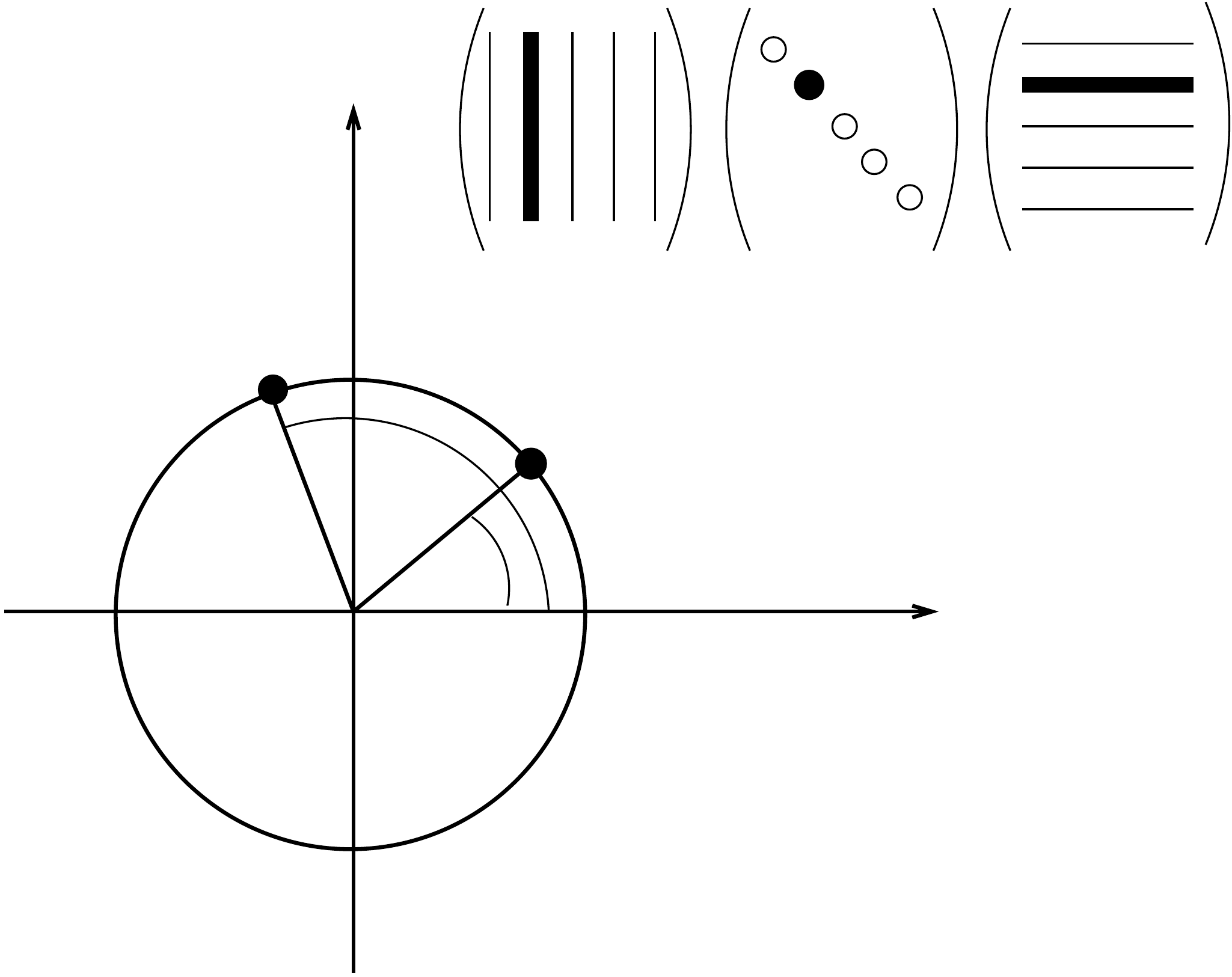} &
        \hspace{-14pt} \resizebox{5.4cm}{!}{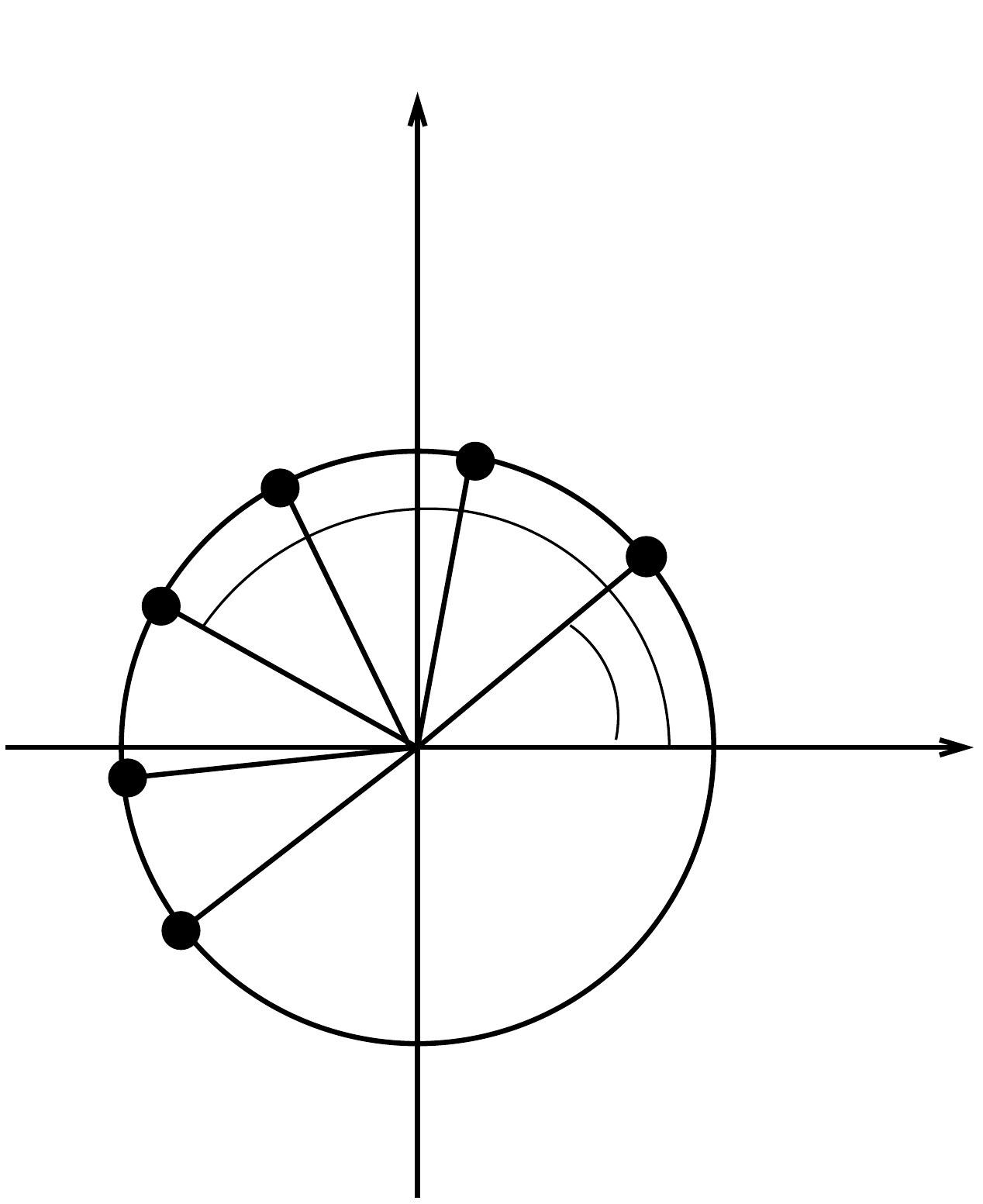} 
    \end{tabular}
    \end{center}
\caption{Training gated sparse coding models is equivalent to detecting rotation 
angles in the invariant subspaces associated with a set of transformations (left), 
and it amounts to detecting multiple applications of the same angle 
when learning from videos (right).}
\label{figure:crosscorrelationandenergy}
\end{figure*}

An important fact about orthogonal matrices is that 
the eigen-decomposition $L=UDU^\mathrm{T}$ is complex, 
where eigenvalues (diagonal of $D$) have absolute value $1$ \cite{HornJohnson}. 
Multiplying by a complex number with absolute value $1$ amounts to performing a rotation in the 
complex plane, as illustrated in Figure \ref{figure:crosscorrelationandenergy} (left). 
Each eigenspace associated with $L$ is also referred to as \emph{invariant subspace} of $L$ 
(as application of $L$ will keep eigenvectors within the subspace). 

Applying an orthogonal warp is thus equivalent to (i) projecting the image onto \emph{filter pairs} 
(the real and imaginary 
parts of each eigenvector), (ii) performing a rotation within each invariant subspace, 
and (iii) projecting back into the image-space. 
In other words, we can decompose an orthogonal transformation into a set of independent, 
2-dimensional rotations.  
The most well-known examples are translations: A 1D-translation matrix contains ones 
along one of its secondary diagonals, and it is zero elsewhere\footnote{To be exactly orthogonal it has to 
contain an additional one in another place, so that it performs a rotation with wrap-around.}. 
The eigenvectors of this matrix are Fourier-components \cite{GrayToeplitzReview}, 
and the rotation in each invariant subspace amounts to a phase-shift of the corresponding Fourier-feature.
This leaves the norm of the projections onto the Fourier-components (the power spectrum of the signal) 
constant, which is a well known property of translation. 


It is interesting to note that the imaginary and real parts of the eigenvectors of a translation matrix 
correspond to 
sine and cosine features, respectively, reflecting the fact that Fourier components naturally come 
in \emph{pairs}. These are commonly referred to as \emph{quadrature pairs} in the literature. 
In the special case of Gabor features, the importance of quadrature pairs is 
that they allow us to detect translations independently of the local content of images
\cite{QianStereoAndMotion, FleetBinocular}. 
However, the property that eigenvectors come in \emph{pairs} is not specific to 
translations. It is shared by all transformations that can be represented by an orthogonal matrix,  
so they can be composed from 2-dimensional rotations. 
\cite{bethge2007} use the term \emph{generalized quadrature pair} to refer to the 
eigen-features of these transformations. 

\subsection{Commuting warps share eigenspaces}
\label{section:commutingwarps}
A central observation to our analysis is that eigenspaces can be 
\emph{shared} among transformations. 
When eigenspaces are shared, then the only way in which two transformations differ, 
is in the angles of rotation within the eigenspaces. 
So shared eigenspaces allow us to represent \emph{multiple transformations with a 
single set of features}. 
An example of a shared eigenspace is the Fourier-basis, which is shared 
among translations. 
This well-known observation follows from the fact that the set of 
all circulant matrices (which are 1-D translation-matrices) 
of the same size have the Fourier-basis as eigen-basis \cite{GrayToeplitzReview}. 
Eigenspaces can be shared between many more transformation not just translation. 
An obvious generalization are 
local translations, which may be considered 
the constituting transformations of natural videos. Another, less obvious generalization is spatial rotation. 
Formally, a set of matrices share eigenvectors if they \emph{commute}\footnote{ 
This can be seen by considering any two matrices $A$ and $B$ with $AB=BA$ and with $\lambda, v$ an 
eigenvalue/eigenvector pair of $B$ with multiplicity one. It holds that 
$
BAv = ABv = \lambda A v.
$
Therefore, $Av$ is also an eigenvector of $B$ with the same eigenvalue.
} \cite{HornJohnson}. 

The importance of commuting transformations for our analysis is that, since these 
transformations share an eigen-basis, they differ only in the angle of rotation in the joint eigenspace. 
As a result, one may \emph{extract} a particular transformation from a given 
image pair ($\bm{x}, \bm{y}$) by recovering the angles of rotation 
between the projections of $\bm{x}$ and $\bm{y}$ onto the eigenspaces. 
For this end, consider the real and complex parts $\bm{v}_\mathrm{R}$ and 
$\bm{v}_\mathrm{I}$ of some eigen-feature $\bm{v}$. That is, 
$\bm{v} = \bm{v}_\mathrm{R} + i \bm{v}_\mathrm{I}$, where $i=\sqrt{-1}$. 
The real and imaginary coordinates of the projection of $\bm{x}$
onto the invariant subspace associated with $\bm{v}$ are given by 
$\bm{v}_\mathrm{R}^\mathrm{T}\bm{x}$ and 
$\bm{v}_\mathrm{I}^\mathrm{T}\bm{x}$, respectively. For the output image, they are 
$\bm{v}_\mathrm{R}^\mathrm{T}\bm{y}$ and 
$\bm{v}_\mathrm{I}^\mathrm{T}\bm{y}$. 

Let $\phi_x$ and $\phi_y$ denote the angles of the projections of $\bm{x}$ and $\bm{y}$
with the real axis in the complex plane. If we normalize the projections to have unit norm, 
then the cosine of the angle between the projections, $\phi_y-\phi_x$, may be written 
$$
\cos(\phi_y-\phi_x)=\cos \phi_y \cos \phi_x + \sin \phi_y \sin \phi_x
$$ 
by trigonometric identity. This is equivalent to computing the inner product between 
two normalized projections (cf. Figure \ref{figure:crosscorrelationandenergy} (left)).
In other words, to estimate the (cosine of) the angle of rotation between the 
projections of $\bm{x}$ and $\bm{y}$, we need to \emph{sum over the product of two filter responses}.  

Note, however, that normalizing each projection to $1$ amounts to dividing by the sum of squared filter 
responses, an operation that is highly unstable if a projection is close to zero. 
Unfortunately, this will be the case, whenever one of the images is almost orthogonal 
to the invariant subspace. This, in turn, means that the rotation angle \emph{cannot 
be recovered from the given image}, because the image is too close to the axis of rotation. 
One may view this as a subspace-generalization of the well-known \emph{aperture problem} 
beyond translation, to the set of orthogonal transformations. 
Normalization would ignore this problem and provide the illusion of a recovered angle 
even when the aperture problem makes the detection of the transformation component impossible. 
In the next section we discuss how one may overcome this problem by rephrasing the problem as a 
\emph{detection} task.

\subsection{Detecting subspace rotations}
\label{section:rotationdetectors}
For each eigenvector, $\bm{v}$, and rotation angle, $\theta$, 
define the complex output image filter
$$
\bm{v}^\theta=\exp(i\theta) \bm{v}
$$
which represents a projection and simultaneous rotation by $\theta$.
This allows us to 
define a {\bf subspace rotation-detector} with preferred angle $\theta$ as follows:
\begin{eqnarray}
\label{equation:detector}
r^\theta = (\bm{v}_R^\mathrm{T}\bm{y}) ({\bm{v}^\theta_R}^\mathrm{T}\bm{x})
+ (\bm{v}_I^\mathrm{T}\bm{y}) ({\bm{v}^\theta_I}^\mathrm{T}\bm{x})
\label{equation:rotationdetector}
\end{eqnarray}
where subscripts $R$ and $I$ denote the real and imaginary part of 
the filters like before.  
Like before, if projections are normalized to length $1$, we have 
$$
r^\theta=\cos \phi_y \cos (\phi_x-\theta) + \sin \phi_y \sin (\phi_x-\theta) = \cos(\phi_y-\phi_x-\theta), 
$$
which is maximal whenever $\phi_y-\phi_x=\theta$, thus when the observed angle of 
rotation, $\phi_y-\phi_x$, 
is equal to the preferred angle of rotation, $\theta$.  
However, like before, normalizing projections is not a good idea because of the subspace 
aperture problem. 
We now show that mapping units are well-suited to detecting subspace rotations, 
if a number of conditions are met.

\subsection{Mapping units as rotation detectors}
If features and data are \emph{contrast normalized},
then the projections will depend only on how well the image pair represents a given subspace rotation.  
The value $r^\theta$, in turn, will depend 
(a) on the transformation (via the subspace angle) 
and 
(b) on the content of the images (via the angle between each image and the invariant subspace). 
Thus, the output of the detector factors in both, the presence of a transformation and 
our ability to discern it.

The fact that $r^\theta$ depends on image content makes it a suboptimal representation 
of transformation.  However, note that $r^\theta$ is a ``conservative'' detector, 
that takes on a large value only if an input image pair $(\bm{x}, \bm{y})$ is compatible 
with its transformation.  
We can therefore define a content-independent representation by \emph{pooling} over multiple 
detectors $r^\theta$ that represent the same transformation but respond to different images. 
Note that computing $r^\theta$ involves summing over the two subspace dimensions, which 
is also a form of pooling (within subspaces).  
Thus, encoding subspace rotations requires two types of pooling. 

If we stack imaginary and real eigenvector pairs for the input and output images, $\bm{v}$ and 
$\bm{v}^\theta$, in matrices $U$ and $V$, respectively, we may define the representation $\bm{t}$ 
of a transformation, given two images $\bm{x}$ and $\bm{y}$, as 
\begin{equation}
\bm{t} = W^\mathrm{T} P \; \big({U}^\mathrm{T}\bm{x}\big) \cdot \big({V}^\mathrm{T}\bm{y}\big)
\label{equation:pooledrotationdetectors}
\end{equation}
where $P$ is a band-diagonal ``within-subspace'' pooling matrix, 
and $W$ is an appropriate ``across-subspace'' pooling matrix. 
Furthermore, the following conditions need to be met: 
(1) Images $\bm{x}$ and $\bm{y}$ are contrast-normalized, 
(2) For each row $\bm{u}_f$ of $U$ there exists $\theta$ such that the corresponding 
row $\bm{v}_f$ of $V$ takes the form $\bm{v}_f=\exp(i\theta)\bm{u}_f$. In other words,
filter pairs are related through rotations only.

Eq. \ref{equation:pooledrotationdetectors} takes exactly the same form as inference in a gated 
sparse coding model (cf., Eq. \ref{equation:factorizationisfiltermatching_inference}), if we absorb the within-subspace 
pooling matrix $P$ into $W$.
\emph{Learning} amounts to identifying both the subspaces and the pooling matrix, 
so training a multi-view feature learning model can be thought of as performing 
multiple simultaneous diagonalizations of a set of transformations. 
When a data-set contains more than one transformation class, learning 
involves partitioning the set of orthogonal warps into commutative subsets and 
simultaneously diagonalizing each subset. 
Note that, in practice, complex filters can be represented by learning two-dimensional 
subspaces in the form of filter pairs. 
It is uncommon, albeit possible, to learn actually complex-valued features in practice.

Diagonalizing a single transformation, $L$, would amount to performing a kind of 
canonical correlations analysis (CCA),  
so learning a multi-view feature learning model 
may be thought of as performing multiple canonical correlation analyzes with tied features. 
Similarly, modeling within-image structure by setting $\bm{x}=\bm{y}$ \cite{mcrbm} would amount 
to learning a PCA mixture with tied weights. 
In the same way that neural networks can be used to implement CCA and PCA up to a linear 
transformation, the result of training a multi-view feature learning model is a simultaneous 
diagonalization only up to a linear transformation. 

It is interesting to note that condition (2) above implies that filters are normalized 
to have the same length. 
Imposing a norm constraint has been a common approach to stabilizing 
learning (eg., \cite{mcrbm, higherordergradientbased, jointImtrans}).
It is also common to apply a sigmoid non-linearity after computing mapping unit activities, 
so that the output of a hidden variable can be interpreted as a probability. 
Pooling over multiple subspaces may, in addition to providing content-independent representations, 
also help deal with edge effects and noise, as well as with the fact that 
learned transformations may not be exactly orthogonal. 

\section{Relation to energy models}
\label{section:energymodels}
By concatenating images $\bm{x}$ and $\bm{y}$, as well as filters $\bm{v}$ and $\bm{v}^\theta$, 
we may approximate the subspace rotation detector (Eq. \ref{equation:detector}) also with the response 
of an energy detector: 
\vspace{-5pt}
\begin{equation}
\begin{split}
r^\theta&=
\big( ({\bm{v}_R}^\mathrm{T}\bm{y}) + 
({\bm{v}^\theta_R}^\mathrm{T}\bm{x})\big)^2 + 
\big( ({\bm{v}_I}^\mathrm{T}\bm{y})+ 
({\bm{v}^\theta_I}^\mathrm{T}\bm{x})\big)^2  
\\
&=
2 \big( ({\bm{v}_R}^\mathrm{T}\bm{y}) ({\bm{v}^\theta_R}^\mathrm{T}\bm{x})
+ ({\bm{v}_I}^\mathrm{T}\bm{y}) ({\bm{v}^\theta_I}^\mathrm{T}\bm{x}) \big)\\
&+ ({\bm{v}_R}^\mathrm{T}\bm{y})^2
+ ({\bm{v}^\theta_R}^\mathrm{T}\bm{x})^2
+ ({\bm{v}_I}^\mathrm{T}\bm{y})^2
+ ({\bm{v}^\theta_I}^\mathrm{T}\bm{x})^2
\end{split}
\label{equation:energydetector}
\end{equation}
Eq. \ref{equation:energydetector} is equivalent to Eq. \ref{equation:detector} up to the four quadratic terms. 
The four quadratic terms are equal to the sum of the squared norms of the projections 
of $\bm{x}$ and $\bm{y}$ onto the invariant subspace. Thus, like the norm of 
the projections, they contribute information about the discernibility of transformations.
This makes the energy response more conservative than the cross-correlation response 
(Eq. \ref{equation:detector}). However, the peak response is still attained only when both 
images reside within the detector's invariant subspace and when their projections are 
rotated by the detectors preferred angle $\theta$.

By pooling over multiple rotation detectors, $r^\theta$, we obtain the equivalent of an 
energy response (Eq.~\ref{equation:energycrosscorrelation}). This shows that 
energy models applied to the concatenation of two images are well-suited to modeling transformations, too.

\subsection{More than two images}
Both energy models and cross-correlation models can be applied to more than two images. 
For gated sparse coding, Eq. \ref{equation:detector} can be modified to contain all cross-terms, 
or all the ones that are deemed relevant (for example, adjacent frames in a ``Markov''-type gating model of a video).
Alternatively, for the energy mechanism, one can compute the square of the concatenation 
of more than two images in place of Eq. \ref{equation:energydetector}. 

\subsubsection{Example: Implementing an energy model via cross-correlation}
\begin{figure}
\begin{center}
\begin{tabular}{cccc}
    Frame: 1 & 2 & 3 & 4 \\
    \scalebox{0.16}[0.16]{\includegraphics[angle=0]{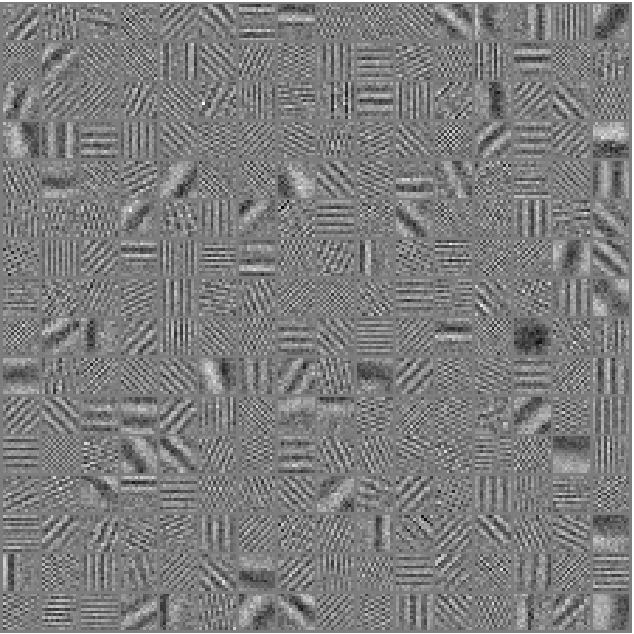}}&
    \scalebox{0.16}[0.16]{\includegraphics[angle=0]{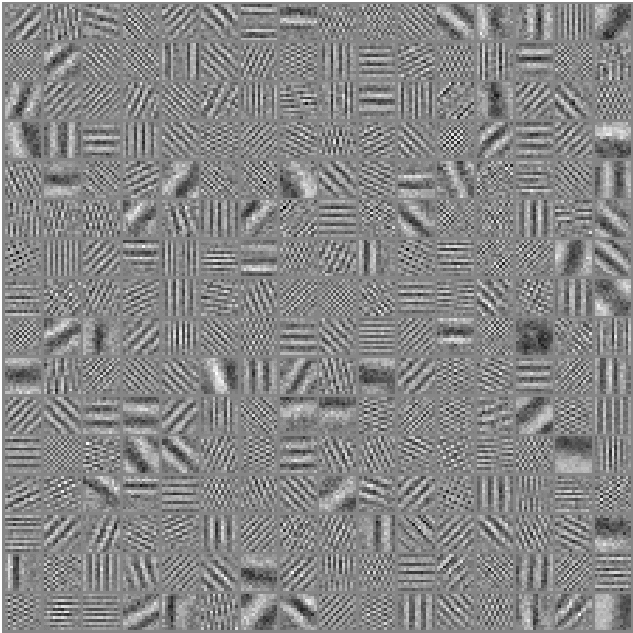}}&
    \scalebox{0.16}[0.16]{\includegraphics[angle=0]{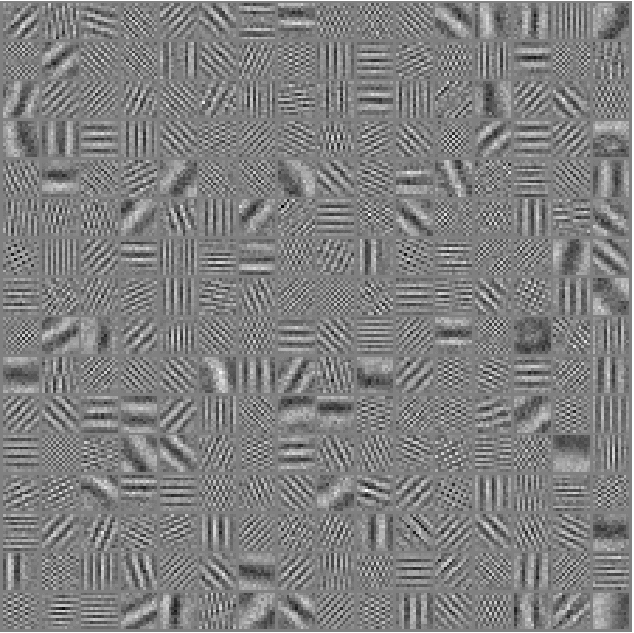}}&
    \scalebox{0.16}[0.16]{\includegraphics[angle=0]{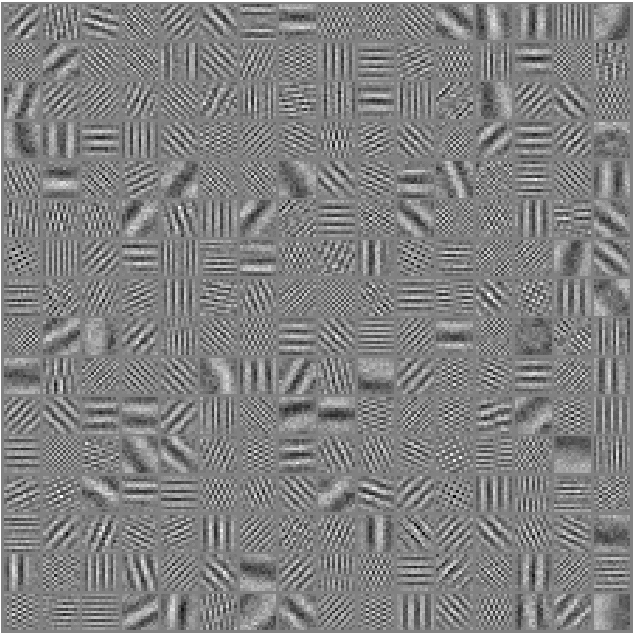}}\\
    5 & 6 & 7 & 8 \\
    \scalebox{0.16}[0.16]{\includegraphics[angle=0]{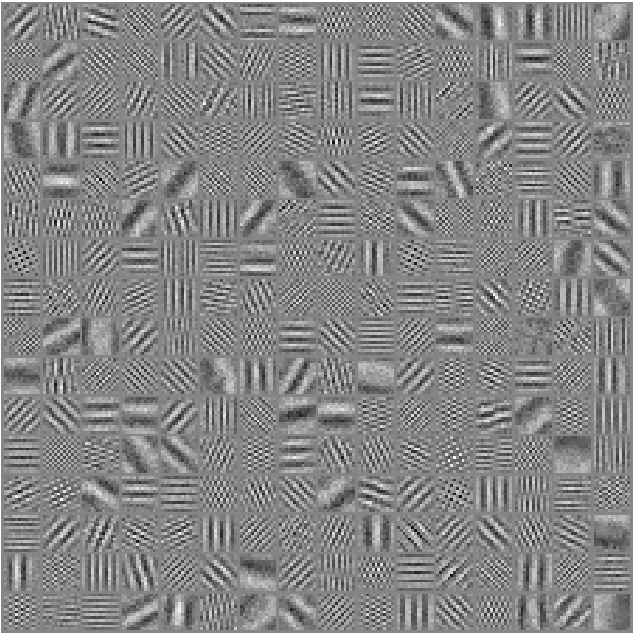}}&
    \scalebox{0.16}[0.16]{\includegraphics[angle=0]{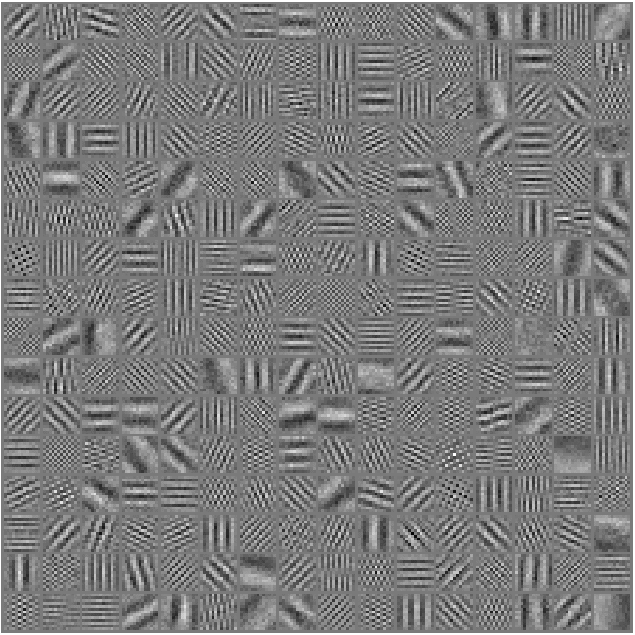}}&
    \scalebox{0.16}[0.16]{\includegraphics[angle=0]{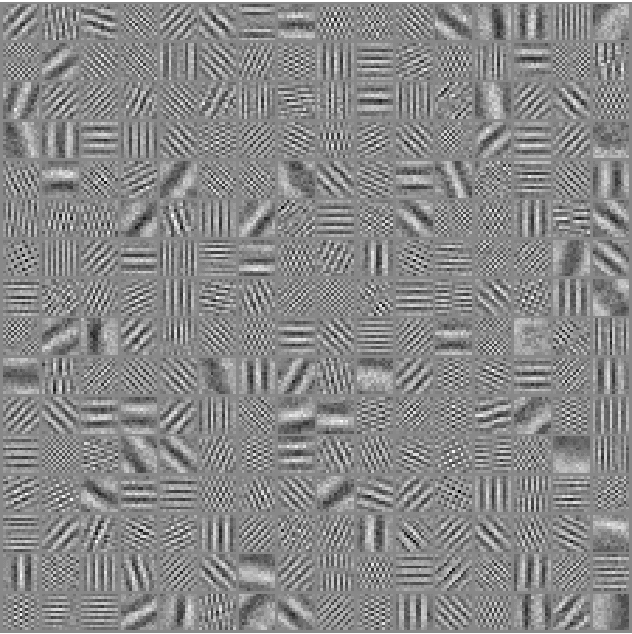}}&
    \scalebox{0.16}[0.16]{\includegraphics[angle=0]{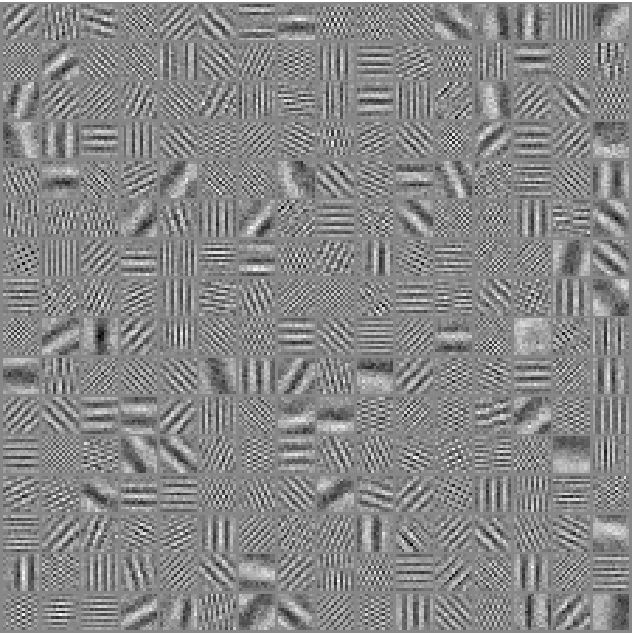}}\\
    9 & 10&& \\
    \scalebox{0.16}[0.16]{\includegraphics[angle=0]{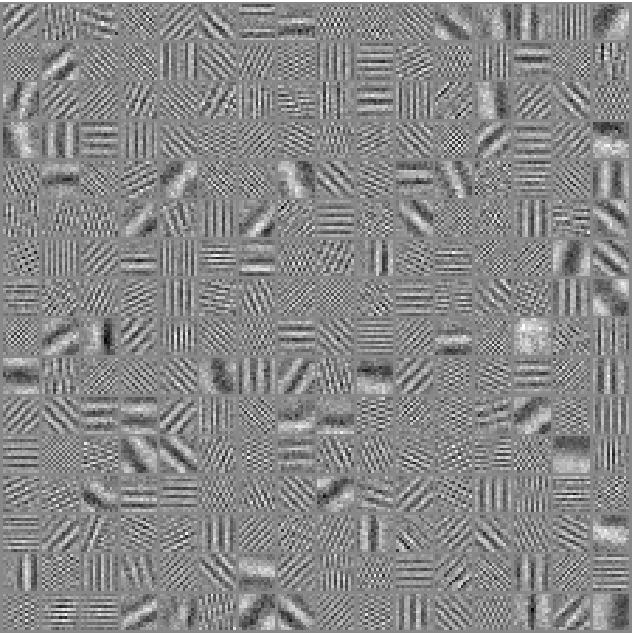}}&
    \scalebox{0.16}[0.16]{\includegraphics[angle=0]{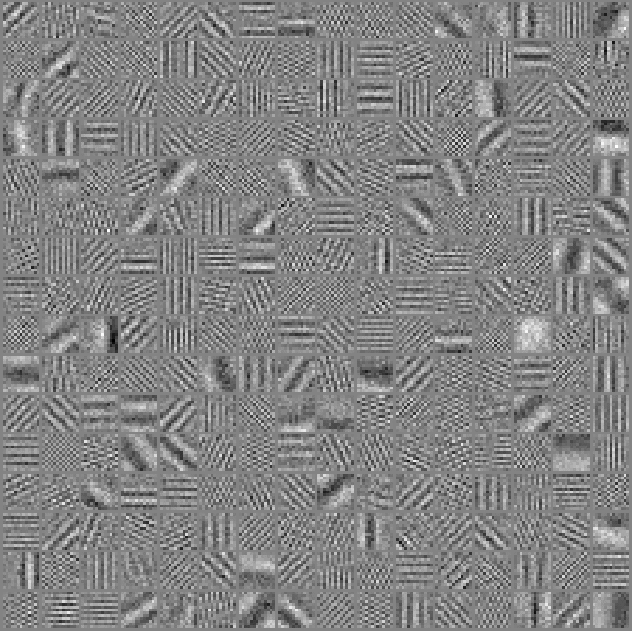}}&&\\
\end{tabular}
\end{center}
\caption{Implementing a \emph{cross-correlation model} via an \emph{energy model} via a \emph{cross-correlation model}. 
Sequence of filters learned from the concatenation of $10$ frames of moving random dots.}
\label{figure:fouriermovie}
\end{figure}

The close relation between energy models and gated sparse coding makes it possible 
to implement one via the other. 
Figure \ref{figure:fouriermovie} shows example filters from an energy model trained 
on concatenated frames from videos showing moving random dots\footnote{The data and an animation of the 
learned spatio-temporal features is available at \url{http://learning.cs.toronto.edu/~rfm/relational}}. 
We trained a gated auto-encoder with $F=256$ factors and $K=128$ 
mapping units, where $\bm{x}=\bm{y}$ is given by the concatenation of $10$ frames. Filters are 
constrained, such that $w^x_{if}=w^y_{if}$. 
Each $10$-frame input shows random dots moving at a constant speed. Speed and direction vary 
across movies. 

Since the gated auto-encoder, a cross-correlation model, multiplies two sets of 
filter responses which are the same, it effectively computes a square, 
and thus implements an energy model. 
In the absence of any within-image structure, all filters learn to represent only across-image correlations.
Thus, as predicted by Eq. \ref{equation:energycrosscorrelation2} the energy model, in turn, implements a 
cross-correlation model.

Figure \ref{figure:fouriermovie} depicts, separately, 
the $10$ sets of $256$ filters corresponding to the $10$ time-frames.
It shows that the model learns spatio-temporal Fourier features which are 
selective for speed, frequency and orientation.

\section{Discussion}
\label{section:discussion}
Given the predominance of correspondence tasks in vision, it seems conceivable that 
the main utility of energy models and complex cells is that they can 
\emph{encode relationships} not (monocular) invariances. 

This suggests that squaring non-linearities, for example, as the transfer function in a
feed-forward network, may be useful, in general, in tasks where relations play a role, 
such as in recognition tasks that involve motion and stereo. 
In the long term, computing squares and/or cross-products could help reduce the requirement 
for large, hand-engineered pipelines, which are currently used for solving 
correspondence problems in tasks like depth inference.  
These typically involve keypoint extraction, descriptor extraction, matching and 
outlier-removal \cite{Hartley2004}. 
A learning based system using complex cells may be able to replace parts of the pipeline 
with a single, homogeneous model that is trained from data. 
This may also help explain how visual cortex may perform a large variety of tasks 
using a single, homogeneous module, which can be trained by a single type of learning mechanism.  

Interestingly, invariant object recognition itself can be viewed as a correspondence 
problem, where the goal is to match an input observation to invariant \emph{templates} 
in memory. \cite{gsm} discuss a variation of 
a gated sparse coding model, which may be considered as an approach to
invariant recognition through modeling \emph{mappings that take images to class labels}. 
The input of the model is an image, the output is an orthogonal encoding of 
a class label, and prediction amounts to marginalizing over the set of possible 
mappings. The graphical model is also equivalent to a set of class-conditional 
manifolds or probability distributions, but inference is feed-forward. 
The model effectively transforms an input into a canonical pose, so that it can 
be matched with a template, which itself represents the object in some canonical pose.
This can help explain the similarity, in general, between filters that allow for 
invariant recognition and those that allow for selective recognition of transformations.  
\cite{gsm} show how ``swirly features'' similar to the rotation features in 
Figures \ref{figure:scrollfiltersx} and \ref{figure:scrollfiltersy} emerge when
learning to perform rotationally invariant recognition. 
\cite{Bergstra10} showed that similar features can emerge in feed-forward recognition 
models that contain squaring non-linearities. 

Most common object recognition systems are somewhat unrealistic in that they are trained 
to recognize single, static views of objects. Real biological systems get to see ``movies'' 
of objects that constantly move around or change their relative pose. It is interesting to 
note that a model that computes squares or cross-products could automatically 
learn to associate object identity with 3-D structure or with articulated motion,
simply by being trained on multiple, concatenated frames. 

Using multiplicative interactions can also be related to analogy making \cite{factoredGBM}.  
It can be argued that analogy making is at the heart of many cognitive 
phenomena \cite{Hofstadter84thecopycat}. An interesting question is, to what degree 
an analogy-making module could be a useful building block in models of 
higher-level cognitive capabilities.
Since gated sparse coding and energy models can be trained with standard, even 
Hebbian-like, learning (cf., Section \ref{section:modelingrelations}), analogy-making 
does not require any uncommon or unusual machinery besides multiplicative interactions. 

Squaring can be approximated using other non-linearities (see, for example,
\cite{zetzsche2005} for a discussion). 
A possible research question is, what type of approximations of computing squares or 
cross-products may be advantageous computationally and/or more plausible biologically. 
Of course, squares could be simulated using a layer of a feed-forward network with 
sigmoid activations \cite{Funahashi}. 
However, the abundance of matching and correspondence tasks in vision may provide some 
inductive bias in favor of genuine multiplicative interactions or squares. 

Another research question is to what degree deviating from exactly commuting 
transformations and exactly orthogonal matrices hampers our ability to learn something useful.
Existing experiments (for example in \cite{imtrans, factoredGBM}) suggest that there 
is some robustness, but there has been no quantitative analysis. 
It is conceivable, that one could pre-process data-points, such that they can be related 
through orthogonal matrices, in order to make them amenable to an energy or cross-correlation model.
Interestingly, it seems that one way to do this, would be by transforming data to 
be high-dimensional and sparse. 

\bibliography{relational}
\bibliographystyle{plain}

\end{document}

%% file: mapping.pdf_t
\begin{picture}(0,0)%
\includegraphics{mapping.pdf}%
\end{picture}%
\setlength{\unitlength}{4144sp}%
\begingroup\makeatletter\ifx\SetFigFont\undefined%
\gdef\SetFigFont#1#2#3#4#5{%
  \reset@font\fontsize{#1}{#2pt}%
  \fontfamily{#3}\fontseries{#4}\fontshape{#5}%
  \selectfont}%
\fi\endgroup%
\begin{picture}(4568,3673)(2671,-3419)
\put(5671,-241){\makebox(0,0)[lb]{\smash{{\SetFigFont{34}{40.8}{\rmdefault}{\mddefault}{\updefault}{\color[rgb]{0,0,0}$z_k$}%
}}}}
\put(2836,-1906){\makebox(0,0)[lb]{\smash{{\SetFigFont{34}{40.8}{\rmdefault}{\mddefault}{\updefault}{\color[rgb]{0,0,0}$x_i$}%
}}}}
\put(6796,-3211){\makebox(0,0)[lb]{\smash{{\SetFigFont{34}{40.8}{\rmdefault}{\mddefault}{\updefault}{\color[rgb]{0,0,0}$y_j$}%
}}}}
\end{picture}%

%% file: bi-partite.pdf_t
\begin{picture}(0,0)%
\includegraphics{bi-partite.pdf}%
\end{picture}%
\setlength{\unitlength}{3947sp}%
\begingroup\makeatletter\ifx\SetFigFont\undefined%
\gdef\SetFigFont#1#2#3#4#5{%
  \reset@font\fontsize{#1}{#2pt}%
  \fontfamily{#3}\fontseries{#4}\fontshape{#5}%
  \selectfont}%
\fi\endgroup%
\begin{picture}(5196,5596)(3703,-4706)
\put(6601,-3736){\makebox(0,0)[lb]{\smash{{\SetFigFont{20}{24.0}{\rmdefault}{\mddefault}{\updefault}{\color[rgb]{0,0,0}$y_j$}%
}}}}
\put(6451,539){\makebox(0,0)[lb]{\smash{{\SetFigFont{25}{30.0}{\rmdefault}{\mddefault}{\updefault}{\color[rgb]{0,0,0}$\bm{z}$}%
}}}}
\put(6601,-886){\makebox(0,0)[lb]{\smash{{\SetFigFont{25}{30.0}{\rmdefault}{\mddefault}{\updefault}{\color[rgb]{0,0,0}$w_{jk}$}%
}}}}
\put(6451,-61){\makebox(0,0)[lb]{\smash{{\SetFigFont{20}{24.0}{\rmdefault}{\mddefault}{\updefault}{\color[rgb]{0,0,0}$z_k$}%
}}}}
\put(6451,-4561){\makebox(0,0)[lb]{\smash{{\SetFigFont{25}{30.0}{\rmdefault}{\mddefault}{\updefault}{\color[rgb]{0,0,0}$\bm{y}$}%
}}}}
\end{picture}%

%% file: autoencoder.pdf_t
\begin{picture}(0,0)%
\includegraphics{autoencoder.pdf}%
\end{picture}%
\setlength{\unitlength}{3947sp}%
\begingroup\makeatletter\ifx\SetFigFont\undefined%
\gdef\SetFigFont#1#2#3#4#5{%
  \reset@font\fontsize{#1}{#2pt}%
  \fontfamily{#3}\fontseries{#4}\fontshape{#5}%
  \selectfont}%
\fi\endgroup%
\begin{picture}(6771,8716)(2653,-9290)
\put(6076,-5011){\makebox(0,0)[lb]{\smash{{\SetFigFont{20}{24.0}{\rmdefault}{\mddefault}{\updefault}{\color[rgb]{0,0,0}$z_k$}%
}}}}
\put(6076,-8311){\makebox(0,0)[lb]{\smash{{\SetFigFont{20}{24.0}{\rmdefault}{\mddefault}{\updefault}{\color[rgb]{0,0,0}$y_j$}%
}}}}
\put(6226,-2761){\makebox(0,0)[lb]{\smash{{\SetFigFont{25}{30.0}{\rmdefault}{\mddefault}{\updefault}{\color[rgb]{0,0,0}$w_{kj}$}%
}}}}
\put(6226,-7186){\makebox(0,0)[lb]{\smash{{\SetFigFont{25}{30.0}{\rmdefault}{\mddefault}{\updefault}{\color[rgb]{0,0,0}$a_{jk}$}%
}}}}
\put(6001,-9136){\makebox(0,0)[lb]{\smash{{\SetFigFont{25}{30.0}{\rmdefault}{\mddefault}{\updefault}{\color[rgb]{0,0,0}$\bm{y}$}%
}}}}
\put(5926,-961){\makebox(0,0)[lb]{\smash{{\SetFigFont{25}{30.0}{\rmdefault}{\mddefault}{\updefault}{\color[rgb]{0,0,0}$\hat{\bm{y}}$}%
}}}}
\put(6001,-1711){\makebox(0,0)[lb]{\smash{{\SetFigFont{20}{24.0}{\rmdefault}{\mddefault}{\updefault}{\color[rgb]{0,0,0}$\hat{y_j}$}%
}}}}
\end{picture}%

%% file: transformation.pdf_t
\begin{picture}(0,0)%
\includegraphics{transformation.pdf}%
\end{picture}%
\setlength{\unitlength}{4144sp}%
\begingroup\makeatletter\ifx\SetFigFont\undefined%
\gdef\SetFigFont#1#2#3#4#5{%
  \reset@font\fontsize{#1}{#2pt}%
  \fontfamily{#3}\fontseries{#4}\fontshape{#5}%
  \selectfont}%
\fi\endgroup%
\begin{picture}(5220,5949)(2431,-7501)
\put(6661,-7261){\makebox(0,0)[lb]{\smash{{\SetFigFont{47}{56.4}{\rmdefault}{\mddefault}{\updefault}{\color[rgb]{0,0,0}$\bm{y}$}%
}}}}
\put(3061,-7216){\makebox(0,0)[lb]{\smash{{\SetFigFont{47}{56.4}{\rmdefault}{\mddefault}{\updefault}{\color[rgb]{0,0,0}$\bm{x}$}%
}}}}
\put(4996,-2131){\makebox(0,0)[lb]{\smash{{\SetFigFont{47}{56.4}{\rmdefault}{\mddefault}{\updefault}{\color[rgb]{0,0,0}$\bm{z}$}%
}}}}
\end{picture}%

%% file: shifts_outer_2.pdf_t
\begin{picture}(0,0)%
\includegraphics{shifts_outer_2.pdf}%
\end{picture}%
\setlength{\unitlength}{4144sp}%
\begingroup\makeatletter\ifx\SetFigFont\undefined%
\gdef\SetFigFont#1#2#3#4#5{%
  \reset@font\fontsize{#1}{#2pt}%
  \fontfamily{#3}\fontseries{#4}\fontshape{#5}%
  \selectfont}%
\fi\endgroup%
\begin{picture}(10728,3753)(328,-7381)
\end{picture}%

%% file: shifts_outer_3.pdf_t
\begin{picture}(0,0)%
\includegraphics{shifts_outer_3.pdf}%
\end{picture}%
\setlength{\unitlength}{4144sp}%
\begingroup\makeatletter\ifx\SetFigFont\undefined%
\gdef\SetFigFont#1#2#3#4#5{%
  \reset@font\fontsize{#1}{#2pt}%
  \fontfamily{#3}\fontseries{#4}\fontshape{#5}%
  \selectfont}%
\fi\endgroup%
\begin{picture}(10728,3711)(328,-7339)
\end{picture}%

%% file: shifts_vecouter_hids.pdf_t
\begin{picture}(0,0)%
\includegraphics{shifts_vecouter_hids.pdf}%
\end{picture}%
\setlength{\unitlength}{4144sp}%
\begingroup\makeatletter\ifx\SetFigFont\undefined%
\gdef\SetFigFont#1#2#3#4#5{%
  \reset@font\fontsize{#1}{#2pt}%
  \fontfamily{#3}\fontseries{#4}\fontshape{#5}%
  \selectfont}%
\fi\endgroup%
\begin{picture}(10683,4242)(2266,-5716)
\put(6301,-1861){\makebox(0,0)[lb]{\smash{{\SetFigFont{29}{34.8}{\rmdefault}{\mddefault}{\updefault}{\color[rgb]{0,0,0}$z_k$}%
}}}}
\put(6301,-4426){\makebox(0,0)[lb]{\smash{{\SetFigFont{29}{34.8}{\rmdefault}{\mddefault}{\updefault}{\color[rgb]{0,0,0}$w_{ijk}$}%
}}}}
\end{picture}%

%% file: gbm1.pdf_t
\begin{picture}(0,0)%
\includegraphics{gbm1.pdf}%
\end{picture}%
\setlength{\unitlength}{3947sp}%
\begingroup\makeatletter\ifx\SetFigFont\undefined%
\gdef\SetFigFont#1#2#3#4#5{%
  \reset@font\fontsize{#1}{#2pt}%
  \fontfamily{#3}\fontseries{#4}\fontshape{#5}%
  \selectfont}%
\fi\endgroup%
\begin{picture}(6096,8154)(3103,-7204)
\put(8776,-3661){\makebox(0,0)[lb]{\smash{{\SetFigFont{20}{24.0}{\rmdefault}{\mddefault}{\updefault}{\color[rgb]{0,0,0}$y_j$}%
}}}}
\put(3376,-3661){\makebox(0,0)[lb]{\smash{{\SetFigFont{20}{24.0}{\rmdefault}{\mddefault}{\updefault}{\color[rgb]{0,0,0}$x_i$}%
}}}}
\put(6301,-61){\makebox(0,0)[lb]{\smash{{\SetFigFont{20}{24.0}{\rmdefault}{\mddefault}{\updefault}{\color[rgb]{0,0,0}$z_k$}%
}}}}
\put(6226,539){\makebox(0,0)[lb]{\smash{{\SetFigFont{29}{34.8}{\rmdefault}{\mddefault}{\updefault}{\color[rgb]{0,0,0}$\bm{z}$}%
}}}}
\put(3301,-7036){\makebox(0,0)[lb]{\smash{{\SetFigFont{29}{34.8}{\rmdefault}{\mddefault}{\updefault}{\color[rgb]{0,0,0}$\bm{x}$}%
}}}}
\put(8776,-7036){\makebox(0,0)[lb]{\smash{{\SetFigFont{29}{34.8}{\rmdefault}{\mddefault}{\updefault}{\color[rgb]{0,0,0}$\bm{y}$}%
}}}}
\end{picture}%

%% file: gbm2.pdf_t
\begin{picture}(0,0)%
\includegraphics{gbm2.pdf}%
\end{picture}%
\setlength{\unitlength}{3947sp}%
\begingroup\makeatletter\ifx\SetFigFont\undefined%
\gdef\SetFigFont#1#2#3#4#5{%
  \reset@font\fontsize{#1}{#2pt}%
  \fontfamily{#3}\fontseries{#4}\fontshape{#5}%
  \selectfont}%
\fi\endgroup%
\begin{picture}(7337,5904)(1562,-4954)
\put(6601,-3736){\makebox(0,0)[lb]{\smash{{\SetFigFont{20}{24.0}{\rmdefault}{\mddefault}{\updefault}{\color[rgb]{0,0,0}$y_j$}%
}}}}
\put(1801,-1561){\makebox(0,0)[lb]{\smash{{\SetFigFont{20}{24.0}{\rmdefault}{\mddefault}{\updefault}{\color[rgb]{0,0,0}$x_i$}%
}}}}
\put(1801,-4786){\makebox(0,0)[lb]{\smash{{\SetFigFont{29}{34.8}{\rmdefault}{\mddefault}{\updefault}{\color[rgb]{0,0,0}$\bm{x}$}%
}}}}
\put(6451,-4786){\makebox(0,0)[lb]{\smash{{\SetFigFont{29}{34.8}{\rmdefault}{\mddefault}{\updefault}{\color[rgb]{0,0,0}$\bm{y}$}%
}}}}
\put(6451,539){\makebox(0,0)[lb]{\smash{{\SetFigFont{29}{34.8}{\rmdefault}{\mddefault}{\updefault}{\color[rgb]{0,0,0}$\bm{z}$}%
}}}}
\put(6451,-61){\makebox(0,0)[lb]{\smash{{\SetFigFont{20}{24.0}{\rmdefault}{\mddefault}{\updefault}{\color[rgb]{0,0,0}$z_k$}%
}}}}
\end{picture}%

%% file: gbm2_y_xz.pdf_t
\begin{picture}(0,0)%
\includegraphics{gbm2_y_xz.pdf}%
\end{picture}%
\setlength{\unitlength}{3947sp}%
\begingroup\makeatletter\ifx\SetFigFont\undefined%
\gdef\SetFigFont#1#2#3#4#5{%
  \reset@font\fontsize{#1}{#2pt}%
  \fontfamily{#3}\fontseries{#4}\fontshape{#5}%
  \selectfont}%
\fi\endgroup%
\begin{picture}(7337,5904)(1562,-4954)
\put(6601,-3736){\makebox(0,0)[lb]{\smash{{\SetFigFont{20}{24.0}{\rmdefault}{\mddefault}{\updefault}{\color[rgb]{0,0,0}$y_j$}%
}}}}
\put(1801,-1561){\makebox(0,0)[lb]{\smash{{\SetFigFont{20}{24.0}{\rmdefault}{\mddefault}{\updefault}{\color[rgb]{0,0,0}$x_i$}%
}}}}
\put(1801,-4786){\makebox(0,0)[lb]{\smash{{\SetFigFont{29}{34.8}{\rmdefault}{\mddefault}{\updefault}{\color[rgb]{0,0,0}$\bm{x}$}%
}}}}
\put(6451,-4786){\makebox(0,0)[lb]{\smash{{\SetFigFont{29}{34.8}{\rmdefault}{\mddefault}{\updefault}{\color[rgb]{0,0,0}$\bm{y}$}%
}}}}
\put(6451,539){\makebox(0,0)[lb]{\smash{{\SetFigFont{29}{34.8}{\rmdefault}{\mddefault}{\updefault}{\color[rgb]{0,0,0}$\bm{z}$}%
}}}}
\put(6451,-61){\makebox(0,0)[lb]{\smash{{\SetFigFont{20}{24.0}{\rmdefault}{\mddefault}{\updefault}{\color[rgb]{0,0,0}$z_k$}%
}}}}
\end{picture}%

%% file: gbm2_z_xy.pdf_t
\begin{picture}(0,0)%
\includegraphics{gbm2_z_xy.pdf}%
\end{picture}%
\setlength{\unitlength}{3947sp}%
\begingroup\makeatletter\ifx\SetFigFont\undefined%
\gdef\SetFigFont#1#2#3#4#5{%
  \reset@font\fontsize{#1}{#2pt}%
  \fontfamily{#3}\fontseries{#4}\fontshape{#5}%
  \selectfont}%
\fi\endgroup%
\begin{picture}(7337,5904)(1562,-4954)
\put(6601,-3736){\makebox(0,0)[lb]{\smash{{\SetFigFont{20}{24.0}{\rmdefault}{\mddefault}{\updefault}{\color[rgb]{0,0,0}$y_j$}%
}}}}
\put(1801,-1561){\makebox(0,0)[lb]{\smash{{\SetFigFont{20}{24.0}{\rmdefault}{\mddefault}{\updefault}{\color[rgb]{0,0,0}$x_i$}%
}}}}
\put(1801,-4786){\makebox(0,0)[lb]{\smash{{\SetFigFont{29}{34.8}{\rmdefault}{\mddefault}{\updefault}{\color[rgb]{0,0,0}$\bm{x}$}%
}}}}
\put(6451,-4786){\makebox(0,0)[lb]{\smash{{\SetFigFont{29}{34.8}{\rmdefault}{\mddefault}{\updefault}{\color[rgb]{0,0,0}$\bm{y}$}%
}}}}
\put(6451,539){\makebox(0,0)[lb]{\smash{{\SetFigFont{29}{34.8}{\rmdefault}{\mddefault}{\updefault}{\color[rgb]{0,0,0}$\bm{z}$}%
}}}}
\put(6451,-61){\makebox(0,0)[lb]{\smash{{\SetFigFont{20}{24.0}{\rmdefault}{\mddefault}{\updefault}{\color[rgb]{0,0,0}$z_k$}%
}}}}
\end{picture}%

%% file: gae.pdf_t
\begin{picture}(0,0)%
\includegraphics{gae.pdf}%
\end{picture}%
\setlength{\unitlength}{3947sp}%
\begingroup\makeatletter\ifx\SetFigFont\undefined%
\gdef\SetFigFont#1#2#3#4#5{%
  \reset@font\fontsize{#1}{#2pt}%
  \fontfamily{#3}\fontseries{#4}\fontshape{#5}%
  \selectfont}%
\fi\endgroup%
\begin{picture}(8805,6094)(2011,-7308)
\put(2851,-4186){\makebox(0,0)[lb]{\smash{{\SetFigFont{20}{24.0}{\rmdefault}{\mddefault}{\updefault}{\color[rgb]{0,0,0}$x_i$}%
}}}}
\put(7651,-4486){\makebox(0,0)[lb]{\smash{{\SetFigFont{20}{24.0}{\rmdefault}{\mddefault}{\updefault}{\color[rgb]{0,0,0}$z_k$}%
}}}}
\put(7726,-7036){\makebox(0,0)[lb]{\smash{{\SetFigFont{20}{24.0}{\rmdefault}{\mddefault}{\updefault}{\color[rgb]{0,0,0}$y_j$}%
}}}}
\put(10801,-7111){\makebox(0,0)[lb]{\smash{{\SetFigFont{25}{30.0}{\rmdefault}{\mddefault}{\updefault}{\color[rgb]{0,0,0}$\bm{y}$}%
}}}}
\put(7726,-1636){\makebox(0,0)[lb]{\smash{{\SetFigFont{20}{24.0}{\rmdefault}{\mddefault}{\updefault}{\color[rgb]{0,0,0}$\hat{y}_j$}%
}}}}
\put(10651,-1711){\makebox(0,0)[lb]{\smash{{\SetFigFont{25}{30.0}{\rmdefault}{\mddefault}{\updefault}{\color[rgb]{0,0,0}$\hat{\bm{y}}$}%
}}}}
\put(2026,-4411){\makebox(0,0)[lb]{\smash{{\SetFigFont{25}{30.0}{\rmdefault}{\mddefault}{\updefault}{\color[rgb]{0,0,0}$\bm{x}$}%
}}}}
\put(9376,-4636){\makebox(0,0)[lb]{\smash{{\SetFigFont{25}{30.0}{\rmdefault}{\mddefault}{\updefault}{\color[rgb]{0,0,0}$\bm{z}$}%
}}}}
\end{picture}%

%% file: factoredtensor.pdf_t
\begin{picture}(0,0)%
\includegraphics{factoredtensor.pdf}%
\end{picture}%
\setlength{\unitlength}{4144sp}%
\begingroup\makeatletter\ifx\SetFigFontNFSS\undefined%
\gdef\SetFigFontNFSS#1#2#3#4#5{%
  \reset@font\fontsize{#1}{#2pt}%
  \fontfamily{#3}\fontseries{#4}\fontshape{#5}%
  \selectfont}%
\fi\endgroup%
\begin{picture}(7819,5992)(1921,-6155)
\put(6166,-6046){\makebox(0,0)[lb]{\smash{{\SetFigFontNFSS{20}{24.0}{\rmdefault}{\mddefault}{\updefault}{\color[rgb]{0,0,0}$w^{y}_{jf}$}%
}}}}
\put(1936,-3571){\makebox(0,0)[lb]{\smash{{\SetFigFontNFSS{20}{24.0}{\rmdefault}{\mddefault}{\updefault}{\color[rgb]{0,0,0}$w^{x}_{if}$}%
}}}}
\put(9226,-3976){\makebox(0,0)[lb]{\smash{{\SetFigFontNFSS{20}{24.0}{\rmdefault}{\mddefault}{\updefault}{\color[rgb]{0,0,0}$w^{z}_{kf}$}%
}}}}
\put(7111,-3301){\makebox(0,0)[lb]{\smash{{\SetFigFontNFSS{20}{24.0}{\rmdefault}{\mddefault}{\updefault}{\color[rgb]{0,0,0}$w_{ijk}$}%
}}}}
\end{picture}%

%% file: factoredgbm.pdf_t
\begin{picture}(0,0)%
\includegraphics{factoredgbm.pdf}%
\end{picture}%
\setlength{\unitlength}{3947sp}%
\begingroup\makeatletter\ifx\SetFigFontNFSS\undefined%
\gdef\SetFigFontNFSS#1#2#3#4#5{%
  \reset@font\fontsize{#1}{#2pt}%
  \fontfamily{#3}\fontseries{#4}\fontshape{#5}%
  \selectfont}%
\fi\endgroup%
\begin{picture}(9396,8526)(1303,-9109)
\put(1351,-8986){\makebox(0,0)[lb]{\smash{{\SetFigFontNFSS{20}{24.0}{\rmdefault}{\mddefault}{\updefault}{\color[rgb]{0,0,0}$\bm{x}$}%
}}}}
\put(10276,-8836){\makebox(0,0)[lb]{\smash{{\SetFigFontNFSS{20}{24.0}{\rmdefault}{\mddefault}{\updefault}{\color[rgb]{0,0,0}$\bm{y}$}%
}}}}
\put(1501,-6511){\makebox(0,0)[lb]{\smash{{\SetFigFontNFSS{20}{24.0}{\rmdefault}{\mddefault}{\updefault}{\color[rgb]{0,0,0}$x_i$}%
}}}}
\put(10276,-6436){\makebox(0,0)[lb]{\smash{{\SetFigFontNFSS{20}{24.0}{\rmdefault}{\mddefault}{\updefault}{\color[rgb]{0,0,0}$y_j$}%
}}}}
\put(6076,-886){\makebox(0,0)[lb]{\smash{{\SetFigFontNFSS{20}{24.0}{\rmdefault}{\mddefault}{\updefault}{\color[rgb]{0,0,0}$\bm{z}$}%
}}}}
\put(6076,-1486){\makebox(0,0)[lb]{\smash{{\SetFigFontNFSS{20}{24.0}{\rmdefault}{\mddefault}{\updefault}{\color[rgb]{0,0,0}$z_k$}%
}}}}
\end{picture}%

%% file: isa.pdf_t
\begin{picture}(0,0)%
\includegraphics{isa.pdf}%
\end{picture}%
\setlength{\unitlength}{3947sp}%
\begingroup\makeatletter\ifx\SetFigFont\undefined%
\gdef\SetFigFont#1#2#3#4#5{%
  \reset@font\fontsize{#1}{#2pt}%
  \fontfamily{#3}\fontseries{#4}\fontshape{#5}%
  \selectfont}%
\fi\endgroup%
\begin{picture}(5763,8983)(4078,-9083)
\put(6826,-1111){\makebox(0,0)[lb]{\smash{{\SetFigFont{20}{24.0}{\rmdefault}{\mddefault}{\updefault}{\color[rgb]{0,0,0}$z_k$}%
}}}}
\put(6901,-511){\makebox(0,0)[lb]{\smash{{\SetFigFont{29}{34.8}{\rmdefault}{\mddefault}{\updefault}{\color[rgb]{0,0,0}$\bm{z}$}%
}}}}
\put(7426,-2611){\makebox(0,0)[lb]{\smash{{\SetFigFont{34}{40.8}{\rmdefault}{\mddefault}{\updefault}{\color[rgb]{0,0,0}$w^z_{kf}$}%
}}}}
\put(5251,-8911){\makebox(0,0)[lb]{\smash{{\SetFigFont{29}{34.8}{\rmdefault}{\mddefault}{\updefault}{\color[rgb]{0,0,0}$\bm{x}$}%
}}}}
\put(7951,-8911){\makebox(0,0)[lb]{\smash{{\SetFigFont{29}{34.8}{\rmdefault}{\mddefault}{\updefault}{\color[rgb]{0,0,0}$\bm{y}$}%
}}}}
\put(5176,-8236){\makebox(0,0)[lb]{\smash{{\SetFigFont{20}{24.0}{\rmdefault}{\mddefault}{\updefault}{\color[rgb]{0,0,0}$x_i$}%
}}}}
\put(7876,-8236){\makebox(0,0)[lb]{\smash{{\SetFigFont{20}{24.0}{\rmdefault}{\mddefault}{\updefault}{\color[rgb]{0,0,0}$y_j$}%
}}}}
\put(9826,-4636){\makebox(0,0)[lb]{\smash{{\SetFigFont{57}{68.4}{\rmdefault}{\mddefault}{\updefault}{\color[rgb]{0,0,0}$(\cdot)^2$}%
}}}}
\put(7876,-7486){\makebox(0,0)[lb]{\smash{{\SetFigFont{34}{40.8}{\rmdefault}{\mddefault}{\updefault}{\color[rgb]{0,0,0}$w^y_{jf}$}%
}}}}
\put(5551,-7561){\makebox(0,0)[lb]{\smash{{\SetFigFont{34}{40.8}{\rmdefault}{\mddefault}{\updefault}{\color[rgb]{0,0,0}$w^x_{if}$}%
}}}}
\end{picture}%

%% file: invar_subspace_1.pdf_t
\begin{picture}(0,0)%
\includegraphics{invar_subspace_1.pdf}%
\end{picture}%
\setlength{\unitlength}{4144sp}%
\begingroup\makeatletter\ifx\SetFigFont\undefined%
\gdef\SetFigFont#1#2#3#4#5{%
  \reset@font\fontsize{#1}{#2pt}%
  \fontfamily{#3}\fontseries{#4}\fontshape{#5}%
  \selectfont}%
\fi\endgroup%
\begin{picture}(9364,7428)(3838,-7609)
\put(6706,-4066){\makebox(0,0)[lb]{\smash{{\SetFigFont{34}{40.8}{\rmdefault}{\mddefault}{\updefault}{\color[rgb]{0,0,0}$\phi_y$}%
}}}}
\put(7156,-4696){\makebox(0,0)[lb]{\smash{{\SetFigFont{34}{40.8}{\rmdefault}{\mddefault}{\updefault}{\color[rgb]{0,0,0}$\phi_x$}%
}}}}
\put(6661,-916){\makebox(0,0)[lb]{\smash{{\SetFigFont{34}{40.8}{\rmdefault}{\mddefault}{\updefault}{\color[rgb]{0,0,0}$\mathrm{Im}$}%
}}}}
\put(10891,-5326){\makebox(0,0)[lb]{\smash{{\SetFigFont{34}{40.8}{\rmdefault}{\mddefault}{\updefault}{\color[rgb]{0,0,0}$\mathrm{Re}$}%
}}}}
\end{picture}%

%% file: invar_subspace_1_multi.pdf_t
\begin{picture}(0,0)%
\includegraphics{invar_subspace_1_multi.pdf}%
\end{picture}%
\setlength{\unitlength}{4144sp}%
\begingroup\makeatletter\ifx\SetFigFont\undefined%
\gdef\SetFigFont#1#2#3#4#5{%
  \reset@font\fontsize{#1}{#2pt}%
  \fontfamily{#3}\fontseries{#4}\fontshape{#5}%
  \selectfont}%
\fi\endgroup%
\begin{picture}(5781,7086)(4063,-7519)
\put(6661,-916){\makebox(0,0)[lb]{\smash{{\SetFigFont{34}{40.8}{\rmdefault}{\mddefault}{\updefault}{\color[rgb]{0,0,0}$\mathrm{Im}$}%
}}}}
\put(5626,-2896){\makebox(0,0)[lb]{\smash{{\SetFigFont{34}{40.8}{\rmdefault}{\mddefault}{\updefault}{\color[rgb]{0,0,0}$\phi_{x_t}$}%
}}}}
\put(7066,-4696){\makebox(0,0)[lb]{\smash{{\SetFigFont{34}{40.8}{\rmdefault}{\mddefault}{\updefault}{\color[rgb]{0,0,0}$\phi_{x_1}$}%
}}}}
\put(9181,-5416){\makebox(0,0)[lb]{\smash{{\SetFigFont{34}{40.8}{\rmdefault}{\mddefault}{\updefault}{\color[rgb]{0,0,0}$\mathrm{Re}$}%
}}}}
\end{picture}%